\documentclass[10pt,twocolumn,letterpaper]{article}

\usepackage{cvpr}              % To produce the CAMERA-READY version
\usepackage{booktabs}
\usepackage{multirow}
\usepackage{colortbl}
\usepackage{xcolor}
\usepackage{float}
\usepackage{makecell}
\usepackage{adjustbox}
\usepackage{amssymb}    % 提供更多数学符号
\usepackage{pifont}     % 提供叮当字体符号
\usepackage{fontawesome5}

\usepackage{subcaption}
\usepackage{amsmath}

\usepackage{algorithm}
\usepackage{algpseudocode}

\definecolor{tfcolor}{gray}{0.93}
\newcommand{\tfc}{\cellcolor{tfcolor}}

\definecolor{cvprblue}{rgb}{0.21,0.49,0.74}
\usepackage[pagebackref,breaklinks,colorlinks,allcolors=cvprblue]{hyperref}

\def\confName{CVPR}
\def\confYear{2026}

\title{WISER: Wider Search, Deeper Thinking, and Adaptive Fusion for Training-Free Zero-Shot Composed Image Retrieval}
\author{
Tianyue Wang$^{1,2}$,
Leigang Qu$^{4}$,
Tianyu Yang$^{1,3}$,
Xiangzhao Hao$^{1,3}$,
Yifan Xu$^{6}$,\\
Haiyun Guo$^{1,3}$\textsuperscript{,\faIcon[regular]{envelope}} ,
Jinqiao Wang$^{1,2,3,5}$\textsuperscript{,\faIcon[regular]{envelope}}  \\
$^1$Foundation Model Research Center, Institute of Automation, Chinese Academy of Sciences \\
$^2$School of Advanced Interdisciplinary Sciences, University of Chinese Academy of Sciences \\
$^3$School of Artificial Intelligence, University of Chinese Academy of Sciences \\
$^4$National University of Singapore 
\quad $^5$Wuhan AI Research
\quad $^6$Minzu University of China \\
{\tt\small wangtianyue25@mails.ucas.ac.cn, leigangqu@gmail.com, 24012460@muc.edu.cn},\\
{\tt\small \{yangtianyu2024, haoxiangzhao2023\}@ia.ac.cn, \{haiyun.guo, jqwang\}@nlpr.ia.ac.cn}
}
\begin{document}
\maketitle
\begin{abstract}

Zero-Shot Composed Image Retrieval (ZS-CIR) aims to retrieve target images given a multimodal query (comprising a reference image and a modification text), without training on annotated triplets. 
Existing methods typically convert the multimodal query into a single modality—either as an edited caption for Text-to-Image retrieval (T2I) or as an edited image for Image-to-Image retrieval (I2I). However, T2I often loses fine-grained visual details, while I2I struggles with complex semantic modifications.
% Existing methods convert multimodal queries into either edited captions or edited images, but each either loses fine-grained visual details or struggles with complex semantic accuracy.
To effectively leverage their complementary strengths under diverse query intents, we propose \textbf{WISER}, a training-free framework that unifies T2I and I2I via a ``retrieve–verify–refine'' pipeline, explicitly modeling \textit{intent awareness} and \textit{uncertainty awareness}. Specifically, WISER first performs \textbf{Wider Search} by generating both edited captions and images for parallel retrieval to broaden the candidate pool. 
Then, it conducts \textbf{Adaptive Fusion} with a verifier to assess retrieval confidence, triggering refinement for uncertain retrievals, and dynamically fusing the dual-path for reliable ones. For uncertain retrievals, WISER generates refinement suggestions through structured self-reflection to guide the next retrieval round toward \textbf{Deeper Thinking}. 
Extensive experiments show the superior performance of WISER across multiple benchmarks, achieving relative improvements of 45\% on CIRCO (mAP@5) and 57\% on CIRR (Recall@1) over existing training-free methods. 
Notably, it even surpasses many training-dependent methods, highlighting its effectiveness under diverse scenarios. 
Code is released at \url{https://github.com/Physicsmile/WISER}.

\end{abstract}

{
\renewcommand{\thefootnote}{} % 让脚注标记为空
\footnotetext{\faIcon[regular]{envelope} Corresponding author.}
}    
\section{Introduction}
\label{sec:intro}
Imagine you aim to find a red leather jacket similar to the one your friend is wearing, but with a hood instead of a collar. You show the jacket to a retrieval system and say, “Add a hood.” This scenario conveys the essence of Composed Image Retrieval (CIR): retrieving a target image that matches a composed query consisting of a reference image and a modification text~\cite{baldrati2022conditioned,chen2020image,lee2021cosmo,vo2019composing,jiang2024cala,long2024cfir,wen2024simple,hao2026trace,hao2025referring}. CIR plays a vital role in applications such as fashion search and product recommendation.
\begin{figure}[t]
  \centering
   \includegraphics[width=0.95\linewidth]{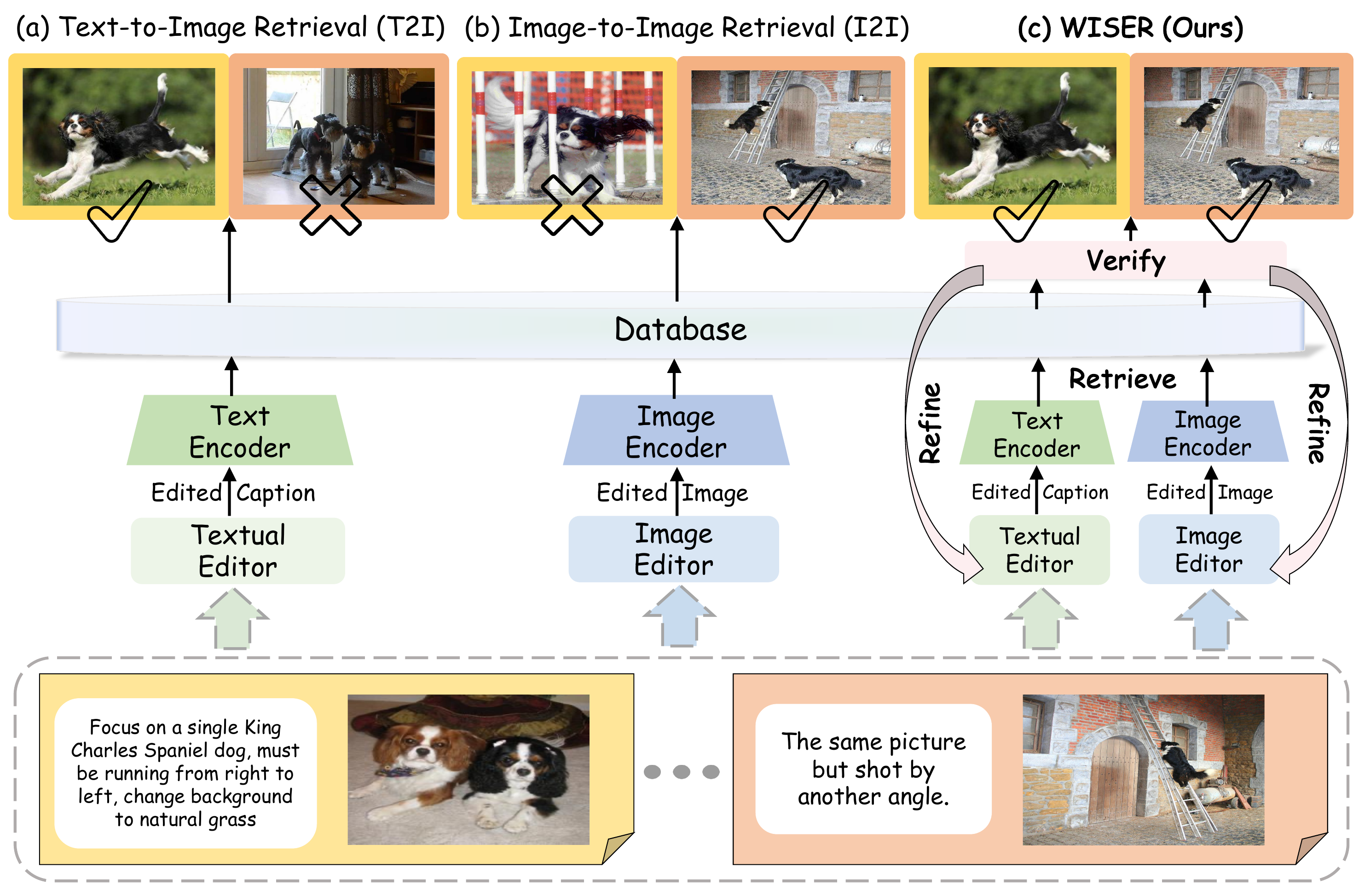}
   \caption{\textbf{Comparison of existing ZS-CIR methods.} (a) T2I may fail to preserve visual details from the reference image, while (b) I2I often struggles with complex modifications. In contrast, (c) WISER successfully adapts to diverse modification intents through a ``retrieve–verify–refine'' pipeline.}
   \label{fig:rw}
   % \vspace{-6pt}
\end{figure}

Most existing CIR methods~\cite{han2017automatic,wang2023cross,vo2019composing,liu2021image,qu2021dynamic, delmas2022artemis,baldrati2022effective} rely on costly annotated triplets, which are labor-intensive and difficult to scale to new domains.
To alleviate this dependency, Zero-Shot CIR (ZS-CIR) has recently been introduced~\cite{baldrati2023zero, saito2023pic2word}, mainly following two paradigms, as shown in Figure \ref{fig:rw}.
The first paradigm leverages the text editor (usually implemented with a textual inversion module~\cite{saito2023pic2word,baldrati2023zero,gu2024language} or an image captioner cascaded with a Large Language Model~\cite{karthik2023vision,yang2024ldre,cheng2025generative}) to transform the composed query into an edited caption, and then performs Text-to-Image retrieval (T2I).
% While effective for complex semantic modifications, 
Although this paradigm shows promise for complex semantic modifications, it often fails to preserve fine-grained visual details from the reference image. 
The second paradigm, in contrast, employs an image editor to produce an edited image by editing the reference image conditioned on the modification text, thus framing CIR as Image-to-Image retrieval (I2I)~\cite{gu2023compodiff,wang2025generative,li2025imagine}. 
It retains visual details better but performs poorly when the query intent is ambiguous or involves complex compositional edits.
% performs poorly when the query intent involves complex compositional edits.
Given the diversity of real-world CIR queries, relying solely on either T2I or I2I is insufficient.
This raises a crucial question: \textit{How can we leverage the complementary strengths of both paradigms to accommodate diverse modification intents?}
However, effectively unifying T2I and I2I for ZS-CIR is non-trivial, due to the following two key challenges: 
(1) \textit{Intent Awareness.} Existing methods often adopt static (e.g., fixed-weight) fusion strategies~\cite{wang2025generative,li2025imagine}, which lack adaptability to varying query intents.
(2) \textit{Uncertainty Awareness.} Current approaches overlook the uncertainty of candidates from each branch, leading to unreliable fusion.

To address these challenges, we propose \textbf{WISER}, a training-free framework to enable Wider Search, Adaptive Fusion, and Deeper Thinking for ZS-CIR. 
WISER generalizes well across domains without additional training, and its modular design is compatible with off-the-shelf models.
Specifically, WISER unifies T2I and I2I paradigms through a ``retrieve--verify--refine'' pipeline, explicitly modeling intent and uncertainty awareness to handle diverse modification intents.
Firstly, we perform \textit{Wider Search} by activating both T2I and I2I pathways in parallel. It uses an editor to produce an edited caption and an edited image, expanding the candidate pool from two complementary views. 
Next, we conduct \textit{Adaptive Fusion} by employing a verifier to evaluate candidates from each branch with confidence scores, determining the reliability of each pathway. For uncertain retrievals, it triggers a refinement process; otherwise, it applies a multi-level fusion strategy that addresses uncertainty awareness at the branch level and intent awareness at the candidate level. 
Finally, we enable \textit{Deeper Thinking} by refining uncertain retrievals via a refiner, which generates improvement suggestions through structured self-reflection to guide the next retrieval round. 
The pipeline iterates until a maximum count is reached.

In summary, our main contributions are as follows:
(1) We are the first to propose a training-free framework for ZS-CIR that adaptively leverages the complementary strengths of both T2I and I2I paradigms.
(2) WISER unifies T2I and I2I through an iterative ``retrieve--verify--refine'' loop, enabling both intent and uncertainty awareness.
(3) WISER demonstrates remarkable superiority and generalization across multiple benchmarks, even surpassing many training-based methods.

\section{Related Work}
\label{sec:related_work}

\subsection{Zero-Shot Composed Image Retrieval}
Composed Image Retrieval (CIR) retrieves target images given a reference image and a modification text as the query. Supervised methods \cite{han2017automatic,wang2023cross,vo2019composing,liu2021image,delmas2022artemis,baldrati2022effective} rely on costly annotated triplets, while Zero-Shot CIR (ZS-CIR) leverages the intrinsic generalization of pretrained vision-language models without depending on any manual triplet annotation~\cite{cohen2022my,saito2023pic2word,baldrati2023zero}. 
Existing ZS-CIR methods mainly follow two paradigms. Text-to-Image retrieval (T2I) approaches
~\cite{saito2023pic2word,baldrati2023zero,gu2024language,karthik2023vision,yang2024ldre,sun2025cotmr}, such as Pic2Word \cite{saito2023pic2word} and SEARLE \cite{baldrati2023zero}, need training-dependent textual inversion modules and suffer from limited visual expressiveness. 
Training-free methods such as CIReVL \cite{karthik2023vision} and CoTMR \cite{sun2025cotmr} utilize Large Language Models (LLMs) or Multimodal Large Language Models (MLLMs) to infer target captions through multimodal reasoning, excelling at semantic changes but potentially losing visual details. 
Image-to-Image retrieval (I2I) approaches, exemplified by CompoDiff \cite{gu2023compodiff}, employ diffusion-based frameworks in CLIP~\cite{radford2021learning} feature space to preserve visual details, though they may struggle with complex edits. 
Recent methods, such as CIG \cite{wang2025generative} and IP-CIR \cite{li2025imagine} edit images via diffusion models~\cite{rombach2022high,podell2023sdxl,zhou2024migc} to augment T2I baselines, yet they still depend on additional training or the manual tuning of fusion hyperparameters. 
In contrast, our work unifies both paradigms in a training-free framework that dynamically balances T2I and I2I, overcoming limitations of static fusion strategies.

\subsection{Vision-Language Models for CIR}
Vision-language models (VLMs) learn aligned representations from large-scale image-text pairs \cite{radford2021learning,li2023blip,li2021align,yu2022coca}, mapping visual and textual inputs into a shared embedding space \cite{chen2023vlp,zhang2024vision,qu2024tiger,qu2025vincie}. The CLIP model \cite{radford2021learning} has demonstrated remarkable zero-shot capabilities across diverse applications. Beyond CIR, VLMs have shown strong performance in zero-shot image classification \cite{cao2023review,lang2025retrieval,lian2025facing,radford2021learning}, visual question answering (VQA) \cite{cai2024forag,naik2023context,zhang2024conditional}, semantic segmentation \cite{cai2025cpsnet}, recommendation systems \cite{fan2023uamc,hu2025bridging}, and social network analysis \cite{cheng2025seeing,cheng2025disentangling,zhou2021survey}. 
Most ZS-CIR methods rely on CLIP variants, which either convert the multimodal query into text embedding space as T2I or into image embedding space as I2I. However, they fail to adaptively harness both strengths based on the specific modification intent.
Our framework overcomes this limitation by adaptively integrating both T2I and I2I pathways through a novel ``retrieve--verify--refine'' pipeline.

% \subsection{Vision-Language Pre-trained Models for CIR}
% Vision-language models (VLMs) have become foundational in multimodal understanding by learning aligned representations from large-scale image-text pairs \cite{radford2021learning,li2023blip,li2021align,yu2022coca}. These models map visual and textual inputs into a shared embedding space, enabling cross-modal tasks without task-specific fine-tuning \cite{chen2023vlp,zhang2024vision}. The CLIP model \cite{radford2021learning}, in particular, has demonstrated remarkable zero-shot capabilities across diverse applications, serving as the backbone for numerous downstream tasks. Beyond Composed Image Retrieval, VLMs have shown strong performance in zero-shot image classification \cite{cao2023review,lang2025retrieval,lian2025facing,radford2021learning}, visual question answering (VQA) \cite{cai2024forag,naik2023context,zhang2024conditional}, semantic segmentation \cite{cai2025cpsnet}, recommendation systems \cite{fan2023uamc,hu2025bridging}, and social network analysis \cite{cheng2025seeing,cheng2025disentangling,zhou2021survey}. In the context of ZS-CIR, most methods build upon CLIP variants of different scales to project images and texts into a unified embedding space for retrieval. Our framework similarly leverages the CLIP backbone for the dual-path retrieval.

\section{Method}
\label{sec:method}

\begin{figure*}[h]
  \centering
  \includegraphics[width=1.0\textwidth]{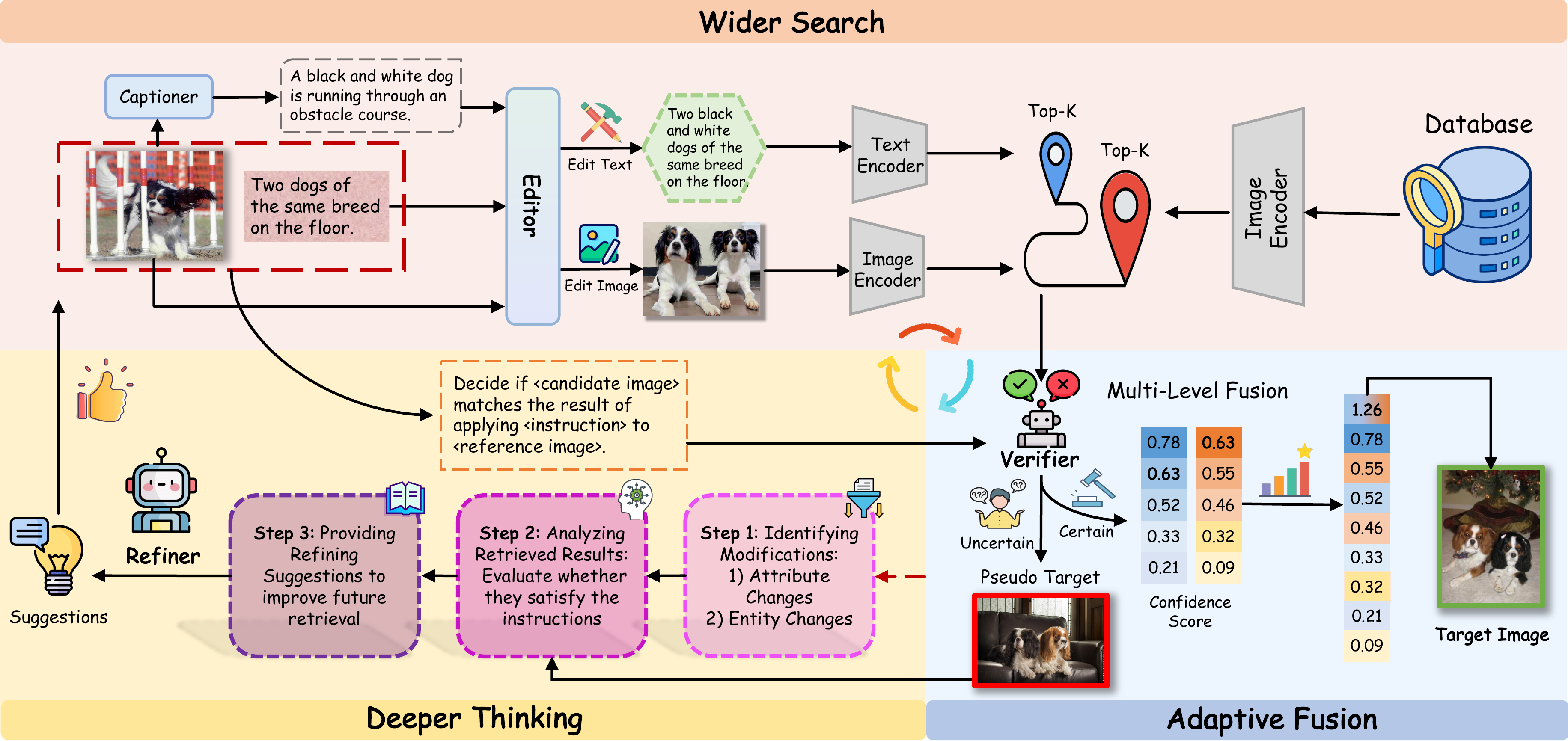}
  \caption{\textbf{Overview of the proposed WISER framework. }(1) Wider Search. We leverage an editor to produce text and image queries for dual-path retrieval, aggregating the top-$K$ results into a unified candidate pool.
  (2) Adaptive Fusion. We employ a verifier to assess the candidates with confidence scores, applying a multi-level fusion strategy for high-confidence results and triggering refinement for low-confidence ones.
  (3) Deeper Thinking. For uncertain retrievals, we leverage a refiner to analyze unmet modifications and then feed targeted suggestions back to the editor, iterating until a predefined limit is reached.}
  \label{fig:pipeline}
\end{figure*}
In this section, we present \textbf{WISER}, a training-free framework for ZS-CIR. As illustrated in Figure~\ref{fig:pipeline}, WISER employs a ``retrieve--verify--refine'' pipeline that unifies T2I and I2I pathways through three core components: Wider Search enables dual-path retrieval (Sec.~\ref{subsec:wider_search}), Adaptive Fusion dynamically evaluates and fuses candidates based on confidence scores (Sec.~\ref{subsec:adaptive_Fusion}), and Deeper Thinking refines uncertain retrievals (Sec.~\ref{subsec:deeper_thinking}). 
Formally, given a composed query including a reference image $I_{\mathrm{ref}}$ and a modification text $T_{\mathrm{mod}}$, the goal of CIR is to retrieve target images from a database $\mathcal{D}$ to match the composed query. In our work, we first employ an \textit{editor} $\mathcal{F}$ to derive an edited image $I_{\text{edit}}$ and an edited caption $C_{\text{edit}}$. They are then encoded into query vectors $q_v$ and $q_t$ using a visual encoder $E_{\text{img}}$ and a text encoder $E_{\text{txt}}$~\cite{radford2021learning}, respectively. The final retrieval is performed by computing the cosine similarity between these query vectors and the candidates in $\mathcal{D}$.

\subsection{Wider Search: Dual-path Retrieval}
\label{subsec:wider_search}
To handle diverse retrieval needs in ZS-CIR, we design \textit{Wider Search}, a strategy that activates both T2I and I2I in parallel to broaden the candidate pool.

\noindent\textbf{Text-to-Image Retrieval (T2I).} We first use a pre-trained captioner to encode the reference image $I_{\text{ref}}$ into a descriptive caption $C_{\text{ref}}$. 
Subsequently, the editor integrates $C_{\text{ref}}$ with the modification text $T_{\text{mod}}$ through its understanding capability, generating an edited caption $C_{\text{edit}}$ that explicitly describes the target image. 
Formally:
\begin{equation}
    C_{\text{edit}} = \mathcal{F}_{\text{txt}}(C_{\text{ref}}, T_{\text{mod}}),
\end{equation}
where $\mathcal{F}_{\text{txt}}$ denotes the textual editing function of the editor. 
This process aims to capture complex semantic modifications specified in the text.

\noindent\textbf{Image-to-Image Retrieval (I2I).} In parallel, the editor leverages its image editing capability to generate an edited image $I_{\text{edit}}$ directly from $I_{\text{ref}}$ and $T_{\text{mod}}$:
\begin{equation}
    I_{\text{edit}} = \mathcal{F}_{\text{img}}(I_{\text{ref}}, T_{\text{mod}}),
\end{equation}
where $\mathcal{F}_{\text{img}}$ denotes the image editing function. 
$I_{\text{edit}}$ retains detailed visual attributes (e.g., texture, style) of the reference image while applying the required modifications.

\noindent\textbf{Dual-path Retrieval.} Given the edited caption $C_{\mathrm{edit}}$ and the edited image $I_{\mathrm{edit}}$, we perform CLIP-based retrieval through two pathways. For each pathway $p \in \{\text{T2I}, \text{I2I}\}$, we retrieve the top-$K$ candidate images:
\begin{equation}
    \mathcal{R}_p = \{ I_{p}^1, I_p^2, \ldots, I_p^K \}.
\end{equation}
We then take the union of both candidate sets to form an expanded candidate pool:
\begin{equation}
\mathcal{R}_{\text{union}} = \mathcal{R}_{\text{T2I}} \cup \mathcal{R}_{\text{I2I}}.
\end{equation}
This operation ensures that all potentially relevant candidates from both T2I and I2I retrievals are considered, thereby improving retrieval recall.

\subsection{Adaptive Fusion: Verification-guided Dual-path Integration}
\label{subsec:adaptive_Fusion}
While \textit{Wider Search} expands the retrieval space by independently producing semantically and visually aligned candidates, a simple union of the dual-path fails to dynamically balance their complementary strengths under diverse query intents.
To address this, we introduce \textit{Adaptive Fusion}, a verification-guided module that explicitly models uncertainty and intent for effective fusion. 
It first verifies candidates to assess the reliability of each pathway. 
If the retrieval is deemed uncertain, it triggers \textit{Deeper Thinking} (Sec.~\ref{subsec:deeper_thinking}) for refinement; otherwise, it dynamically integrates candidates from both paths via a multi-level fusion strategy.

\noindent\textbf{Verification-based Scoring.} For each candidate $I_{p}^k \in \mathcal{R}_\text{union}$, we construct a triplet $(I_{\text{ref}}, T_{\text{mod}}, I_{p}^k)$ and evaluate it using a VLM-based \textit{verifier} $\Phi$ to assess whether the modification intent is faithfully reflected in the candidate image.
The verifier is asked to answer a binary question: \textit{“Decide if the candidate image matches the result of applying the instruction to the reference image.”}
% \begin{quote}
% \textit{“Decide if the candidate image matches the result of applying the instruction to the reference image.”}
% \end{quote}
Let $\ell^{(\mathrm{yes})}_{p,k}$ and $\ell^{(\mathrm{no})}_{p,k}$ denote the logits corresponding to the answer of ``yes'' and ``no'', respectively:
\begin{equation}
    \{\ell^{(\mathrm{yes})}_{p,k}, \ \ell^{(\mathrm{no})}_{p,k}\}
    = 
    \Phi(I_{\mathrm{ref}}, T_{\mathrm{mod}}, I_{p}^k).
\end{equation}
The confidence score $c_p^k$ that $I_p^k$ is a correct modification is computed as:
\begin{equation}
    c_{p}^k = 
    \frac{\exp(\ell^{(\mathrm{yes})}_{p,k})}
         {\exp(\ell^{(\mathrm{yes})}_{p,k}) + \exp(\ell^{(\mathrm{no})}_{p,k})}.
    \label{eq:conf_score_clean}
\end{equation}
A higher confidence score indicates stronger alignment between the candidate and the query intent.

\noindent\textbf{Multi-Level Fusion Strategy.}
Our fusion strategy operates at two levels to address uncertainty and intent awareness, respectively.

\noindent\textit{Branch-Level Uncertainty Awareness.} We first assess the reliability of each pathway by identifying the most promising candidate from each pathway as the pseudo-target $I^{*}_{p}$:
\begin{equation}
    I^{*}_{p} = \arg \max_{k} c_{p}^k, \quad r_p = \max_{k} c_{p}^k,
\end{equation}
where $r_p$ represents the reliability score of pathway $p$. If $\min(r_\text{T2I}, r_\text{I2I})<\tau$ (i.e., either pathway demonstrates uncertainty), the corresponding pseudo-target is forwarded to \textit{Deeper Thinking} (Sec.~\ref{subsec:deeper_thinking}) for refinement rather than proceeding to fusion.

\noindent\textit{Candidate-Level Intent Awareness.} For reliable retrievals, we dynamically integrate candidates through intent-aware fusion. We first compute a fused confidence score that aggregates evidence from both pathways:
\begin{equation}
\text{c}_{\text{fused}}^k = c^{k}_\text{T2I} + c_{\text{I2I}}^k.
\end{equation}
T2I typically yields higher confidence for semantic-heavy edits, while I2I excels when visual details are key. When both aspects matter, candidates strong in both semantics and visuals gain higher fused scores and superior ranks.
% Each candidate image $I_{p}^k \in \mathcal{R}_{\mathrm{union}}$ may appear in one or both retrieval sets. For candidates that do not appear in a particular path, the corresponding confidence score is set to 0 by default.
Then, we sort $\mathcal{R}_{\mathrm{union}}$ using a lexicographical order defined by:
\begin{equation}
    \Psi(I^k) = \left(-\text{c}_{\text{fused}}^k,\ -\max\left( c_{\text{T2I}}^k, c_{\text{I2I}}^k \right),\ -c_{\text{T2I}}^k \right).
\end{equation}
% The sorting key primary ranking by fused score captures overall alignment, while tie-breakers (maximum single-path confidence and T2I confidence) provide nuanced intent-aware disambiguation.
The primary sorting key is the fused score, which captures overall intent alignment. For tie-breaking, it uses maximum single-path confidence and T2I confidence to provide nuanced, intent-aware disambiguation.
Unlike conventional fusion strategies that rely on fixed weights or additional training modules, we dynamically integrate the two retrieval pathways with both intent and uncertainty awareness. 
% effectively leveraging their complementary strengths.

% It ensures that candidates are primarily ranked by their fused score in descending order. For ties, the maximum single-path confidence and then the confidence score from T2I are used as tie-breakers, guaranteeing a unique ordering.
% This fusion process effectively balances the contributions of both retrieval pathways, allowing semantically rich and visually coherent candidates to dominate the final ranking.

\subsection{Deeper Thinking: Structured Self-reflection for Uncertain Retrievals}
\label{subsec:deeper_thinking}
For uncertain retrievals from Adaptive Fusion ($r_{\text{p}}<\tau$), we invoke \textit{Deeper Thinking}, a refinement module that improves the quality of the corresponding edited caption and edited image by analyzing modification failures and generating targeted suggestions.
The refinement is driven by an LLM-based \textit{refiner} that performs a three-step analysis:

\noindent\textbf{Step 1: Identifying Modifications.}  
Given the caption of the reference image $C_{\text{ref}}$ and the modification text $T_{\text{mod}}$, the refiner deeply analyzes the intended changes and generates structured modification phrases. Specifically, it identifies two types of modifications: (1) Attribute Changes: If the modification involves changing characteristics of an entity in $I_{\text{ref}}$, the refiner specifies the change; (2) Entity Additions/Deletions: If the modification adds or removes an entity, the refiner specifies the operation.

\noindent\textbf{Step 2: Analyzing Retrieved Results.}  
We first obtain the caption of the pseudo-target $I_{\text{p}}^{*}$. It is then compared with the modification phrases from Step 1 to determine whether the retrieved image satisfies the user's instruction. This comparison reveals which aspects of the modification are missed or incorrectly applied.

\noindent\textbf{Step 3: Providing Refining Suggestions.} 
For any unmet modifications, the refiner proposes concise, targeted suggestions to improve future retrieval: For T2I, it generates a textual suggestion to enhance the edited caption.
For I2I, it provides visual guidance to improve the edited image generation. Otherwise, the refinement loop is terminated, and the current retrieved images are kept as the final output.

The suggestions are then concatenated with the modification text and fed back to the editor $\mathcal{F}$ to regenerate refined $C_{\text{edit}}$ or $I_{\text{edit}}$, continuing the ``retrieve–verify–refine'' cycle. This process iterates until the maximum number of iterations $N$ is reached.
Deeper Thinking mimics human-like introspection, enabling WISER to self-reflect and adapt to complex or ambiguous queries without any training.

\section{Experiments}
\label{sec:experiments}
We conduct extensive experiments to investigate the effectiveness of our proposed method.
\begin{table*}[h]
\centering
\caption{\textbf{Comparison with the state-of-the-art methods on the test sets of CIRCO and CIRR.} The best results are in \textbf{bold}, while the second-best results are \underline{underlined}. `-' indicates results not reported in the original paper. Our method is highlighted in grey.}
\small
\setlength{\tabcolsep}{3pt}
\begin{adjustbox}{max width=\textwidth}
% \begin{tabular}{c c c | cccc | cccc | ccc}
\begin{tabular}{c|l c | cccc | cccc | ccc}
\toprule
\multirow{2}{*}[-4pt]{\textbf{Backbone}} &
\multirow{2}{*}[-4pt]{\textbf{Method}} &
\multirow{2}{*}[-6pt]{\shortstack{\textbf{Training}\\[-2pt]\textbf{-free}}} &
\multicolumn{4}{c|}{\textbf{CIRCO}} &
\multicolumn{7}{c}{\textbf{CIRR}} \\ 
\cmidrule(lr){4-7} \cmidrule(lr){8-14}
 & & & \multicolumn{4}{c|}{mAP@k} & \multicolumn{4}{c|}{Recall@k} & \multicolumn{3}{c}{Recall$_{sub}$@k} \\ 
 & & & $k=5$ & $k=10$ & $k=25$ & $k=50$ & $k=1$ & $k=5$ & $k=10$ & $k=50$ & $k=1$ & $k=2$ & $k=3$ \\ 
\midrule

\multirow{8}{*}{ViT-B/32} 
& PALAVRA (ECCV'22) & \ding{55} & 4.61 & 5.32 & 6.33 & 6.80 & 16.62 & 43.49 & 58.51 & 83.95 & 41.61 & 65.30 & 80.94 \\
& SEARLE (ICCV'23) & \ding{55} & 9.35 & 9.94 & 11.13 & 11.84 & 24.00 & 53.42 & 66.82 & 89.78 & 54.89 & 76.60 & 88.19 \\
& CIReVL (ICLR’24) & \checkmark & 14.94 & 15.42 & 17.00 & 17.82 & 23.94 & 52.51 & 66.00 & 86.95 & 60.17 & 80.05 & 90.19 \\
& LDRE (SIGIR’24) & \checkmark & 17.96 & 18.32 & 20.21 & 21.11 & 25.69 & 55.13 & 69.04 & 89.90 & 60.53 & 80.65 & 90.70 \\
& OSrCIR (CVPR’25) & \checkmark & 18.04 & 19.17 & 20.94 & 21.85 & 25.42 & 54.54 & 68.19 & - & 62.31 & 80.86 & 91.13 \\
& AutoCIR (KDD'25) & \checkmark  & 18.82 & 19.41 & 21.38 & 22.32 & 30.53 & 59.42 & 72.19 & \underline{91.47} & 65.11 & 84.02 & \textbf{92.70}\\
& CoTMR (ICCV’25) & \checkmark  & \underline{22.23} & \underline{22.78} & \underline{24.68} & \underline{25.74} & 
\underline{31.50} & \underline{60.80} & \underline{73.04} & 91.06 & \underline{66.61} & \underline{84.50} & \underline{92.55} \\
& \tfc \textbf{WISER (Ours)} & \tfc \checkmark & \tfc \textbf{32.23} & \tfc \textbf{33.18} & \tfc \textbf{34.82} & \tfc \textbf{35.35} &
\tfc \textbf{49.45} & \tfc \textbf{76.55} & \tfc \textbf{85.21} & \tfc \textbf{93.81} & \tfc \textbf{77.30} & \tfc \textbf{88.63} & \tfc 92.27 \\
\midrule

\multirow{10}{*}{ViT-L/14}
& SEARLE (ICCV'23) & \ding{55} & 11.68 & 12.73 & 14.33 & 15.12 & 24.24 & 52.48 & 66.29 & 88.84 & 53.76 & 75.01 & 88.19 \\
& LinCIR (CVPR’24) & \ding{55} & 12.59 & 13.58 & 15.00 & 15.85 & 25.04 & 53.25 & 66.68 &  - & 57.11 & 77.37 & 88.89 \\
& MOA (SIGIR’25) & \ding{55} & 15.30 & 17.10 & 18.50 & 19.30 & 27.10 & 56.50 & 69.20 &  90.00 & - & - & - \\
& HIT (ICCV’25) & \ding{55} &  15.50 & 16.70 & 18.90 & 19.90 & 27.90 & 57.60 & 70.50 &  90.40 & - & - & - \\
& CIReVL (ICLR’24) & \checkmark & 18.57 & 19.01 & 20.89 & 21.80 & 24.55 & 52.31 & 64.92 & 86.34 & 59.54 & 79.88 & 89.69 \\
& LDRE (SIGIR’24) & \checkmark & 23.35 & 24.03 & 26.44 & 27.50 & 26.53 & 55.57 & 67.54 & 88.50 & 60.43 & 80.31 & 89.90 \\
& IP-CIR (CVPR’25) & \checkmark & 26.43 & 27.41 & 29.87 & 31.07 & 29.76 & 58.82 & 71.21 & 90.41 & 62.48 & 81.64 & 90.89 \\
& OSrCIR (CVPR’25) & \checkmark & 23.87 & 25.33 & 27.84 & 28.97 & 29.45 & 57.68 & 69.86 & - & 62.12 & 81.92 & 91.10 \\
& AutoCIR (KDD'25) & \checkmark  & 24.05 & 25.14 & 27.35 & 28.36 & 31.81 & 61.95 & 73.86 & 92.07 & 67.21 & 84.89 & \underline{93.13} \\
& CoTMR (ICCV’25) & \checkmark  & \underline{27.61} & \underline{28.22} & \underline{30.61} & \underline{31.70} & \underline{35.02} & \underline{64.75} & \underline{76.18} & \underline{92.51} & \underline{69.39} & \underline{85.75} & \textbf{93.33} \\
& \tfc \textbf{WISER (Ours)} & \tfc \checkmark & \tfc \textbf{35.10} & \tfc \textbf{36.30} & \tfc \textbf{38.46} & \tfc \textbf{39.15} &
\tfc \textbf{49.23}  & \tfc \textbf{76.72}  & \tfc \textbf{85.11}  & \tfc \textbf{94.17}  & \tfc \textbf{77.81} & \tfc \textbf{88.89} & \tfc 92.77 \\
\midrule

\multirow{7}{*}{ViT-G/14}
& LinCIR (CVPR’24) & \ding{55} & 19.71 & 21.01 & 23.13 & 24.18 & 35.25 & 64.72 & 76.05 &  - & 63.35 & 82.22 & 91.98 \\
& CIReVL (ICLR’24) & \checkmark & 26.77 & 27.59 & 29.96 & 31.03 & 34.65 & 64.29 & 75.06 & 91.66 & 67.95 & 84.87 & 93.21 \\
& LDRE (SIGIR’24) & \checkmark & 31.12 & 32.24 & 34.95 & 36.03 & 36.15 & 66.39 & 77.25 & 93.95 & 68.82 & 85.66 & {93.76} \\
& IP-CIR (CVPR’25) & \checkmark & 32.75 & 34.26 & 36.86 & 38.03 & 39.25 & 70.07 & 80.00 & 94.89 & 69.95 & 86.87 & \textbf{94.22} \\
& OSrCIR (CVPR’25) & \checkmark & 30.47 & 31.14 & 35.03 & 36.59 & 37.26 & 67.25 & 77.33 & - & 69.22 & 85.28 & 93.55 \\
& CoTMR (ICCV’25) & \checkmark  & \underline{32.23} & \underline{32.72} & \underline{35.60} & \underline{36.83} & \underline{36.36} & \underline{67.52} & \underline{77.82} & \underline{93.99} & \underline{71.19} & \underline{86.34} & \underline{93.87} \\
& \tfc \textbf{WISER (Ours)} & \tfc \checkmark & \tfc \textbf{36.53} & \tfc \textbf{38.14} & \tfc \textbf{40.46} & \tfc \textbf{41.26} &
\tfc \textbf{49.54} & \tfc \textbf{77.40} & \tfc \textbf{85.76} & \tfc \textbf{94.17} & \tfc \textbf{78.10} & \tfc \textbf{89.06} & \tfc 92.68 \\
\bottomrule
\end{tabular}
\end{adjustbox}
\label{tab:complete_circo}
\end{table*}

\begin{table*}[h]
\centering
\small
\caption{\textbf{Comparison with the state-of-the-art methods on the validation set of Fashion-IQ.} The best results are in \textbf{bold}, while the second-best results are \underline{underlined}. Our method is highlighted in grey.}
\begin{tabular}{c|l c | cc | cc | cc | cc}
\toprule
\multirow{2}{*}[-4pt]{\textbf{Backbone}} &
\multirow{2}{*}[-4pt]{\textbf{Method}} &
\multirow{2}{*}[-4pt]{\shortstack{\textbf{Training}\\[-2pt]\textbf{-free}}} &
\multicolumn{2}{c|}{\textbf{Shirt}} &
\multicolumn{2}{c|}{\textbf{Dress}} &
\multicolumn{2}{c|}{\textbf{Toptee}} &
\multicolumn{2}{c}{\textbf{Avg.}} \\
\cmidrule(lr){4-5} \cmidrule(lr){6-7} \cmidrule(lr){8-9} \cmidrule(lr){10-11}
 & & & R@10 & R@50 & R@10 & R@50 & R@10 & R@50 & R@10 & R@50 \\
\midrule

\multirow{8}{*}{ViT-B/32} 
& PALAVRA (ECCV'22) & \ding{55} & 21.49 & 37.05 & 17.25 & 35.94 & 20.55 & 38.76 & 19.76 & 37.25\\
& SEARLE (ICCV'23) & \ding{55} & 24.44 & 41.61 & 18.54 & 39.51 & 25.70 & 46.46 & 22.89 & 42.53 \\
& CIReVL (ICLR’24) & \checkmark & 28.36 & 47.84 & 25.29 & 46.36 & 31.21 & 53.85 & 28.29 & 49.35 \\
& LDRE (SIGIR’24) & \checkmark & 27.38 & 46.27 & 19.97 & 41.84 & 27.07 & 48.78 & 24.81 & 45.63\\
& OSrCIR (CVPR’25) & \checkmark & 31.16 & 51.13 & 29.35 & 50.37 & 36.51 & 58.71 & 32.34 & 53.40\\
& AutoCIR (KDD'25) & \checkmark & 32.43 & 51.67 & 26.52 & 46.36 & 33.96 & 56.09 & 30.97 & 51.37 \\
& CoTMR (ICCV’25) & \checkmark & \underline{33.42} & \underline{53.93} & \underline{31.09} & \underline{54.54} & \underline{38.40} & \underline{61.14} & \underline{34.30} & \underline{56.54} \\
& \tfc \textbf{WISER (Ours)} & \tfc \checkmark & \tfc \textbf{41.51} & \tfc \textbf{56.92} & \tfc \textbf{37.48} & \tfc \textbf{56.97} & \tfc \textbf{46.97} & \tfc \textbf{62.32} & \tfc \textbf{41.99} & \tfc \textbf{58.74}\\
\midrule

\multirow{9}{*}{ViT-L/14} 
& SEARLE (ICCV'23) & \ding{55} & 26.89 & 45.58 & 20.48 & 43.13 & 29.32 & 49.97 & 25.56 & 46.23 \\
& LinCIR (CVPR’24) & \ding{55} & 29.10 & 46.81 & 20.92 & 42.44 & 28.81 & 50.18 & 26.28 & 46.49 \\
& MOA (SIGIR’25) & \ding{55} & 31.90 & 50.70 & 25.20 & 48.50 & 33.20 & 54.80 & 30.10 & 51.30 \\
& HIT (ICCV’25) & \ding{55} & 32.40 & 51.20 & 25.60 & 47.10 & 32.80 & 54.70 & 30.30 & 51.00 \\
& CIReVL (ICLR’24) & \checkmark & 29.49 & 47.40 & 24.79 & 44.76 & 31.36 & 53.65 & 28.55 & 48.57 \\
& LDRE (SIGIR’24) & \checkmark & 31.04 & 51.22 & 22.93 & 46.76 & 31.57 & 53.64 & 28.51 & 50.54 \\
& OSrCIR (CVPR’25) & \checkmark & 33.17 & 52.03 & 29.70 & 51.81 & 36.92 & 59.27 & 33.26 & 54.37\\
& AutoCIR (KDD'25) & \checkmark & 34.00 & 53.43 & 24.94 & 45.81 & 33.10 & {55.58} & 30.68 & 51.60 \\
& CoTMR (ICCV’25) & \checkmark & \underline{35.43} & \underline{54.91} & \underline{31.18} & \underline{55.04} & \underline{38.55} & \underline{61.33} & \underline{35.05} & \underline{57.09} \\
& \tfc \textbf{WISER (Ours)} & \tfc \checkmark & \tfc \textbf{43.13} & \tfc \textbf{57.85} & \tfc \textbf{38.42} & \tfc \textbf{55.92} & \tfc \textbf{45.39} & \tfc \textbf{61.50} & \tfc \textbf{42.17} & \tfc \textbf{58.51} \\
\midrule

\multirow{7}{*}{ViT-G/14} 
& LinCIR (CVPR’24) & \ding{55} & \textbf{46.76} & \textbf{65.11} & {38.08} & \textbf{60.88} & \textbf{50.48} & \textbf{71.09} & \textbf{45.11} & \textbf{65.69} \\
& CIReVL (ICLR’24) & \checkmark & 33.71 & 51.42 & 27.07 & 49.53 & 35.80 & 56.14 & 32.19 & 52.36\\
& LDRE (SIGIR’24) & \checkmark & 35.94 & 58.58 & 26.11 & 51.12 & 35.42 & 56.67 & 32.49 & 55.46 \\
& OSrCIR (CVPR’25) & \checkmark & 38.65 & 54.71 & 33.02 & 54.78 & 41.04 & 61.83 & 37.57 & 57.11\\
& AutoCIR (KDD'25) & \checkmark & 36.36 & 55.84 & 26.18 & 47.69 & 37.28 & 60.38 & 33.27 & 54.63 \\
& CoTMR (ICCV’25) & \checkmark & 38.32 & \underline{62.24} & 34.51 & {57.36} & 41.90 & 64.30 & 38.25 & 61.32 \\
& \tfc \textbf{WISER (Ours)} & \tfc \checkmark & \tfc \underline{45.98} & \tfc {61.58} & \tfc \textbf{38.57} & \tfc \underline{58.35} & \tfc \underline{49.21} & \tfc \underline{66.96} & \tfc \underline{44.59} & \tfc \underline{62.30} \\
\bottomrule
\end{tabular}
\label{tab:complete_fashioniq}
\end{table*}

\subsection{Experimental Setup}
\noindent\textbf{Datasets.}
We evaluate our proposed WISER on three ZS-CIR benchmarks: \textbf{Fashion-IQ}~\cite{wu2021fashion}, \textbf{CIRR}~\cite{liu2021image} and \textbf{CIRCO} \cite{baldrati2023zero}. 
Fashion-IQ focuses on fashion retrieval with three subcategories (Dress, Shirt, Toptee), 
% containing 30,135 triplets and 77,683 images.
We use Recall@10, Recall@50 as the evaluation metric, and their mean across the three categories.
CIRR is the first natural image dataset designed specifically for CIR, built from real-life images in an open domain. We report Recall@k $(k \in \{1, 5, 10, 50\})$ and Recall$_{\text{Subset}}$@k ($k \in \{1, 2, 3\}$) following the benchmark protocol.
CIRCO is built from the COCO 2017 unlabeled set and is the first CIR dataset providing multiple ground truths per query. It contains 123{,}403 images in the database and 800 test queries. Following prior works, we report mAP@k ($k \in \{5,10,25,50\}$) as the evaluation metric.

\noindent\textbf{Implementation Details.}
We adopt BAGEL \cite{deng2025emerging} as the editor, Qwen2.5-VL-7b \cite{bai2025qwen2} as the verifier, and GPT-4o \cite{achiam2023gpt} as the refiner.
Following previous methods \cite{karthik2023vision}, we use the pre-trained BLIP-2 \cite{li2023blip} as our captioner.
For the retrieval model, we experiment with different CLIP variants, including ViT-B/32, ViT-L/14, and ViT-G/14 CLIP from OpenCLIP \cite{ilharco2021openclip}.
The retrieval candidate pool size $K$ is set to 50, and the reliability threshold $\tau$ is 0.7.
For uncertain cases, the refinement iteration is set to one round by default.
All experiments are implemented using PyTorch~\cite{paszke2019pytorch} on a single NVIDIA H20 GPU, and the results are reported on official validation or test splits.

\noindent\textbf{Baselines.}
We compare WISER with a wide range of representative ZS-CIR baselines.  
Specifically, PALAVRA \cite{cohen2022my}, SEARLE \cite{baldrati2023zero}, LinCIR \cite{gu2024language}, MOA~\cite{li2025rethinking} and HIT~\cite{li2025hierarchy} are textual inversion approaches originally designed or adapted for the ZS-CIR task.  
We also include recent training-free methods such as CIReVL \cite{karthik2023vision}, LDRE \cite{yang2024ldre}, OSrCIR \cite{tang2025reason}, AutoCIR \cite{cheng2025generative}, and CoTMR \cite{sun2025cotmr}, which leverage LLMs or MLLMs without additional training. We also compare with IP-CIR \cite{li2025imagine} under the training-free setting reported in paper, which fuses T2I and I2I pathways using a fixed weight.

\subsection{ZS-CIR Benchmark Comparisons}
\label{subsec:comparison}
We conduct comprehensive comparisons between WISER and state-of-the-art ZS-CIR methods on three benchmark datasets: CIRCO, CIRR, and Fashion-IQ. The results are summarized in Table~\ref{tab:complete_circo} and Table~\ref{tab:complete_fashioniq}. 

\noindent\textbf{CIRCO.} As shown in the left section of Table~\ref{tab:complete_circo}, WISER achieves strong performance on the CIRCO dataset. 
% which features multiple ground truths per query and clean annotations.
Key observations are as follows:
(1) WISER outperforms both training-free and training-based methods across all CLIP backbones. It achieves a relative improvement of 44.98\% in mAP@5 over CoTMR with ViT-B/32, and surpasses LinCIR by 22.51\% in mAP@5 with ViT-L/14.
(2) WISER consistently achieves superior mAP across all $k$ by integrating candidates from both T2I and I2I pathways, expanding the search space and improving retrieval under the multi-target nature of CIRCO.

\noindent\textbf{CIRR.} The right section of Table~\ref{tab:complete_circo} presents results on the challenging CIRR dataset.
%which is known for high noise and weak correlation between reference and target images. 
Key observations are as follows:
(1) WISER delivers strong performance. With ViT-B/32, it achieves a relative improvement of 56.98\% in Recall@1 over the best baseline, demonstrating robust capability in handling diverse and ambiguous modification intents.
(2) In subset recall evaluation, which requires retrieving the correct image from six curated samples, WISER achieves 77.30\% Recall$_{\text{sub}}$@1 with ViT-B/32, surpassing the second-best method by 10.69\%. The improvement is more pronounced for smaller $k$, confirming its effectiveness in ranking the most relevant images at the top.
(3) While some baselines show performance saturation with larger backbones, WISER exhibits consistent gains, with ViT-G/14 achieving a Recall@1 of 49.54\%, far exceeding other methods. This scalability validates the generalizability of our framework.

\noindent\textbf{Fashion-IQ.} Table~\ref{tab:complete_fashioniq} reports results on the Fashion-IQ validation set, which focuses on fine-grained attribute modifications in the fashion domain. 
% This dataset features strong correlations between reference and target images, requiring models to understand semantic attributes while preserving the original style structure. 
Key observations are as follows:
(1) WISER outperforms both training-free and training-dependent baselines using ViT-B/32 and ViT-L/14 by leveraging complementary information from the T2I and I2I pathways, balancing semantic accuracy with visual consistency.
(2) With ViT-G/14, WISER achieves comparable or superior performance to training-based methods such as LinCIR despite lacking their training advantage, further confirming the strong generalization and effectiveness of our approach.

\begin{table}[t]
\centering
\caption{\textbf{Ablation study on the core components of WISER.} ``AVG'' denotes average and ``ADA'' denotes Adaptive Fusion.}
\renewcommand{\arraystretch}{1.2}
\resizebox{\linewidth}{!}{
\begin{tabular}{cc|c|c|cc|cccc}
\toprule
\multicolumn{2}{c|}{\textbf{Wider Search}} & 
\multirow{2}{*}{\makecell[c]{\textbf{Deeper}\\\textbf{Thinking}}} & 
\multirow{2}{*}{\makecell[c]{\textbf{Fusion}}} & 
\multicolumn{2}{c|}{\textbf{Fashion-IQ-Avg}} &
\multicolumn{4}{c}{\textbf{CIRCO}} \\
\cmidrule(lr){1-2} \cmidrule(lr){5-6} \cmidrule(lr){7-10}
\textbf{T2I} & \textbf{I2I} &  &  & \textbf{R@10} & \textbf{R@50} & \textbf{mAP@5} & \textbf{mAP@10} & \textbf{mAP@25} & \textbf{mAP@50} \\
\midrule
- & \checkmark & - & - & 22.65  & 38.84 & 7.00 & 7.46 & 8.40 & 8.95 \\
- & \checkmark & \checkmark & - & 23.58 & 40.10 & 7.57 & 8.05 & 9.04 & 9.62 \\
\checkmark & - & - & - & 28.59  & 49.18 & 17.28 & 17.94 & 19.64 & 20.51 \\
\checkmark & - & \checkmark & - & 29.22 & 49.94 & 17.64 & 18.30 & 20.01 & 20.94 \\
\checkmark & \checkmark & - & AVG & 33.40 & 52.92 & 13.53 & 14.30 & 15.96 & 16.77 \\
\checkmark & \checkmark & - & ADA & 40.83  & 57.86 & 31.32 & 32.08 & 33.72 & 34.24 \\
\checkmark & \checkmark & \checkmark & ADA & 41.99 & 58.74 & 32.23 & 33.18 & 34.82 & 35.35 \\
\bottomrule
\end{tabular}
}
\label{tab:ablation_fashioniq_circo}
\end{table}

% To validate the effectiveness of each component in WISER, we conduct systematic ablation studies. All experiments are performed using the ViT-B/32 backbone on the Fashion-IQ and CIRCO datasets.
\subsection{Ablation Study}
In this section, we conduct experiments on Fashion-IQ and CIRCO with ViT-B/32 to provide an in-depth analysis of WISER, including the effectiveness of its core components, compatibility with various modules, hyperparameter sensitivity, and computational efficiency.

\noindent\textbf{Effectiveness of Core Components.}
Table \ref{tab:ablation_fashioniq_circo} presents the ablation results for the core components of our method. 
We observe the following key findings:
(1) Relying solely on either T2I or I2I yields limited performance, with I2I being even weaker than T2I, primarily due to its difficulty in accurately interpreting semantic modifications. These results confirm \textit{the inherent limitations of the single retrieval paradigm under diverse CIR demands.}
(2) Simply merging the dual-path with fixed weights at the similarity level ('AVG' row) can even degrade performance compared to single-path retrieval.
% , showing that naive merging fails to leverage complementary strengths.
We further compare various fixed fusion strategies with WISER (see Figure \ref{fig:fixed_weight} in supplementary materials). The results show that WISER significantly outperforms all fixed-weight combinations, highlighting \textit{the limitations of manual weight tuning} and verifying the effectiveness of our Adaptive Fusion in leveraging complementary strengths.
(3) Introducing Deeper Thinking on top of Adaptive Fusion leads to further consistent improvements. Moreover, applying Deeper Thinking to single-path baselines also helps.
These results indicate that \textit{iterative analysis effectively corrects retrieval errors.}
(4) Notably, T2I (using BAGEL) performs comparably to the T2I-based baseline CIReVL (e.g., 28.59\% vs. 28.27\% R@10 on Fashion-IQ), indicating no decisive advantage from the editor alone. 
Our full framework achieves substantial improvements, nearly doubling the single-path performance, which verifies that the gains are attributable to our novel design \textbf{rather than merely relying on more powerful models.}

\begin{figure*}[h]
  \centering
   \includegraphics[width=1.0\textwidth]{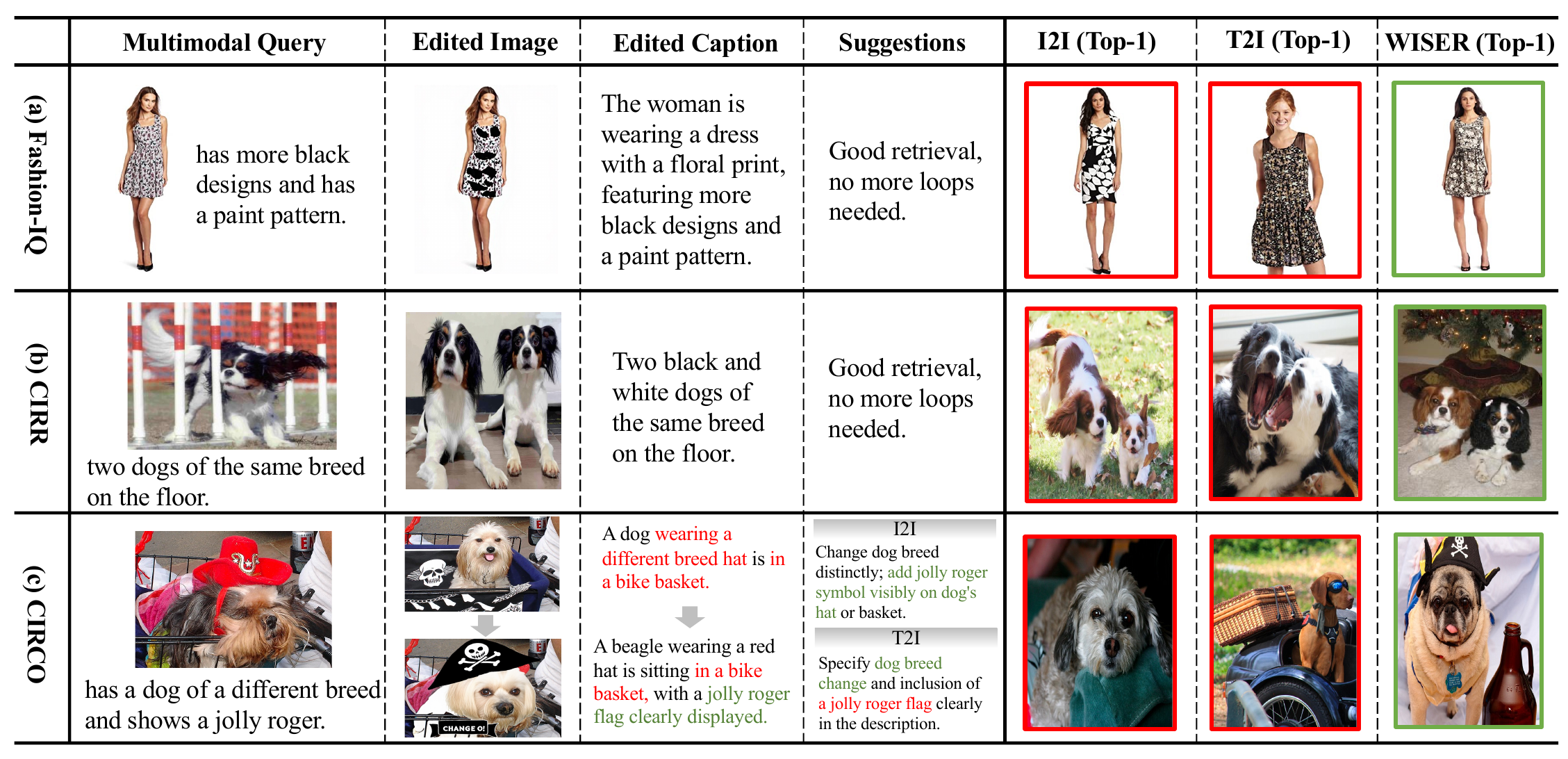}

   \caption{\textbf{Qualitative results on (a) Fashion-IQ, (b) CIRR, and (c) CIRCO datasets.} Red indicates wrong, green represents correct and the gray arrow points to refined results.}
   \label{fig:visualize}
\end{figure*}

\begin{table}[t]
\centering
\caption{\textbf{Ablation study on the choice of editor, verifier, and refiner models on the CIRCO dataset, evaluated with mAP@k.}}
\resizebox{\linewidth}{!}{
\begin{tabular}{c|c|c|ccccc}
\toprule
\textbf{Editor} & \textbf{Verifier} & \textbf{Refiner} & \textbf{k=5} & \textbf{k=10} & \textbf{k=25} & \textbf{k=50} \\
\midrule
\multirow{4}{*}{Bagel~\cite{deng2025emerging}} & \multirow{4}{*}{Qwen2.5-VL-7B~\cite{bai2025qwen2}} & Qwen-Turbo~\cite{bai2023qwen} & 32.80 & 33.56 & 35.21 & 35.79 \\
 & & GPT-3.5-Turbo~\cite{brown2020language} & 32.57 & 33.49 & 35.13 & 35.63 \\
 & & GPT-4o-Mini~\cite{brown2020language} & 32.21 & 33.06 & 34.74 & 35.28 \\
 & & GPT-4o~\cite{achiam2023gpt} & 32.23 & 33.18 & 34.82 & 35.35 \\
\midrule
\multirow{4}{*}{Bagel~\cite{deng2025emerging}} & Qwen2-VL-7B~\cite{wang2024qwen2} & \multirow{4}{*}{GPT-4o~\cite{achiam2023gpt}} & 25.50 & 26.26 & 28.41 & 29.12 \\
 & Qwen2.5-VL-3B~\cite{bai2025qwen2} & & 27.50 & 28.45 & 30.16 & 30.79 \\
 & Qwen2.5-VL-7B~\cite{bai2025qwen2} & & 32.23 & 33.18 & 34.82 & 35.35 \\
 & Qwen2.5-VL-32B~\cite{bai2025qwen2} & & 31.69 & 32.26 & 34.22 & 34.70 \\
\midrule
GPT4o + OmniGen2~\cite{wu2025omnigen2} & \multirow{4}{*}{Qwen2.5-VL-7B~\cite{bai2025qwen2}} & \multirow{4}{*}{GPT-4o~\cite{achiam2023gpt}} & 31.18 & 32.16 & 33.82 & 34.34 \\
GPT4o + Step1X-Edit~\cite{liu2025step1x} &  &  & 31.91 & 33.02 & 34.92 & 35.46 \\
GPT4o + Bagel-Edit~\cite{deng2025emerging} &  &  & 32.21 & 33.25 & 34.99 & 35.51 \\
Bagel~\cite{deng2025emerging} &  &  & 32.23 & 33.18 & 34.82 & 35.35 \\
\bottomrule
\end{tabular}
}
\label{tab:module_ablation}
\end{table}

\noindent\textbf{Compatibility with Various Modules.}
We analyze the effects of different editor, verifier, and refiner modules in Table \ref{tab:module_ablation}.
(1) Refiner: Different LLMs achieve comparable performance, indicating that WISER is robust to the choice of the specific LLM for refinement.
(2) Verifier: Performance generally improves with model scale, though the 32B variant shows slight degradation, probably due to overthinking. Despite this, all verifier configurations achieve high performance, surpassing the best baseline.
(3) Editor: While performance varies across editing models, our framework maintains satisfactory results.
These findings demonstrate WISER's plug-and-play nature, effectively working with various off-the-shelf models.

\begin{figure}[h]
  \centering
   \includegraphics[width=1.0\linewidth]{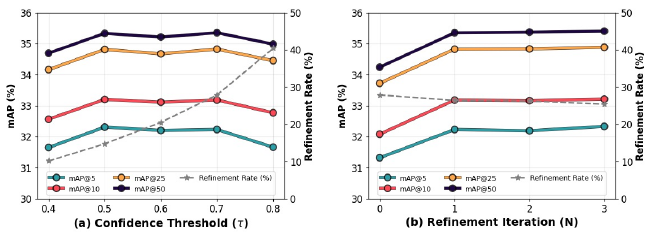}
   \caption{\textbf{Sensitivity analysis on confidence threshold $\tau$ and refinement iteration N on CIRCO.}}
   \label{fig:threshold}
    \vspace{-5pt}
\end{figure}

\noindent\textbf{Parameter Sensitivity of Threshold and Iteration.}
We analyze the impact of the confidence threshold $\tau$ in Deeper Thinking (Figure~\ref{fig:threshold}, left). As $\tau$ increases from 0.4 to 0.8, mAP first rises then slightly declines, indicating that overly low thresholds miss improvement opportunities while overly high ones trigger unnecessary refinement. Performance remains consistently strong when $\tau$ lies between 0.5 and 0.7, allowing flexible trade-offs between efficiency and effectiveness.
We further examine refinement iterations $N$ (Figure~\ref{fig:threshold}, right). All metrics show rapid improvement in the first iteration, with only marginal gains in subsequent rounds. This demonstrates that one iteration generally achieves stable performance, while additional iterations yield diminishing returns at higher computational cost.

\noindent\textbf{Efficiency Analysis.}
WISER's computational overhead is justified by substantial performance gains, as it requires only 0.5 GPU hours per 1\% of improvement. 
Moreover, its plug-and-play architecture enables a flexible trade-off between efficiency and effectiveness, since any component can be replaced with a faster variant. 
As the most expensive component, Deeper Thinking is triggered only for low-confidence cases, which are determined by the threshold $\tau$. 
For reasonable values of $\tau$, the refinement rate remains below 30\% (see Figure~\ref{fig:threshold}), allowing direct retrieval for most queries. 
Notably, WISER already surpasses prior work in a single retrieval round. 
Therefore, users can flexibly adjust $\tau$ and $N$ based on practical needs.

% since users can replace any component with faster variants to meet specific requirements.
% Notably, WISER outperforms prior works in single-round retrieval, with further gains available at marginal cost through optional iterations. Users can thus flexibly tune $\tau$ and the iteration count $N$ to suit deployment requirements.
% While WISER's default configuration involves computationally intensive operations such as image editing and iterative refinement, its plug-and-play architecture allows for the trade-off between efficiency and effectiveness. Users can replace any component with faster variants to meet specific requirements.

\subsection{Qualitative Results}
Figure~\ref{fig:visualize} presents qualitative examples from WISER across three datasets. Each case shows the composed query, edited image, edited caption, refinement suggestions, and top-1 results from T2I-only, I2I-only, and WISER.
In Figure~\ref{fig:visualize}(a), the query requires preserving the reference's style while adding pattern changes. T2I captures semantic changes but loses global style cues, while I2I maintains appearance but misunderstands target patterns. WISER successfully reranks the target image to top-1 by prioritizing candidates with higher overall consistency.
Figure~\ref{fig:visualize}(b) involves fine-grained breed identification and background replacement. T2I over-abstracts breed details, while I2I struggles with precise retrieval. By jointly verifying breed cues and background changes, WISER dynamically balances visual fidelity and semantic precision to achieve top-1 retrieval.
Figure~\ref{fig:visualize}(c) demonstrates handling ambiguous modifications. Initial T2I and I2I both fail—T2I misinterprets key intent while I2I misplaces visual elements. 
As a result, WISER activates Deeper Thinking to refine both paths: enhancing semantic recovery for T2I and correcting visual placement for I2I. In the subsequent retrieval, the improved edited image and caption enable correct top-1 identification, showcasing WISER's robustness through self-reflection.
More details and failure cases are provided in \cref{sec:more_qualitative} of the supplementary material.
\section{Conclusion}
\label{sec:conclusion}
In this work, we introduced WISER, a novel training-free framework for ZS-CIR. To address the diverse nature of real-world user intents, WISER leverages the complementary strengths of both Text-to-Image retrieval (T2I) and Image-to-Image retrieval (I2I) paradigms through a proposed ``retrieve–verify–refine'' pipeline. 
Extensive experiments across multiple benchmarks demonstrate that WISER achieves new state-of-the-art performance, significantly outperforming previous training-free methods and even surpassing many training methods. 
This remarkable improvement confirms WISER's strong superiority and generalization capability across diverse modification intents.
We believe WISER represents a significant step towards more intelligent and adaptable CIR systems. Future work will explore extending the ``retrieve–verify–refine'' pipeline to a wider range of retrieval tasks and further optimizing the framework's efficiency for real-time applications.

\section*{Acknowledgements}
This work was supported in part by the National Key R\&D Program of China (No. 2022ZD0160601), Beijing Natural Science Foundation (L252035), and the National Natural Science Foundation of China under Grant 62276260.

{
    \small
    \bibliographystyle{ieeenat_fullname}
    \bibliography{main}
}

\clearpage
\setcounter{page}{1}
\maketitlesupplementary

% \section{Complete Ablation Study}

\section{More Ablation Study}
\begin{figure}[t]
  \centering
   \includegraphics[scale=0.25]{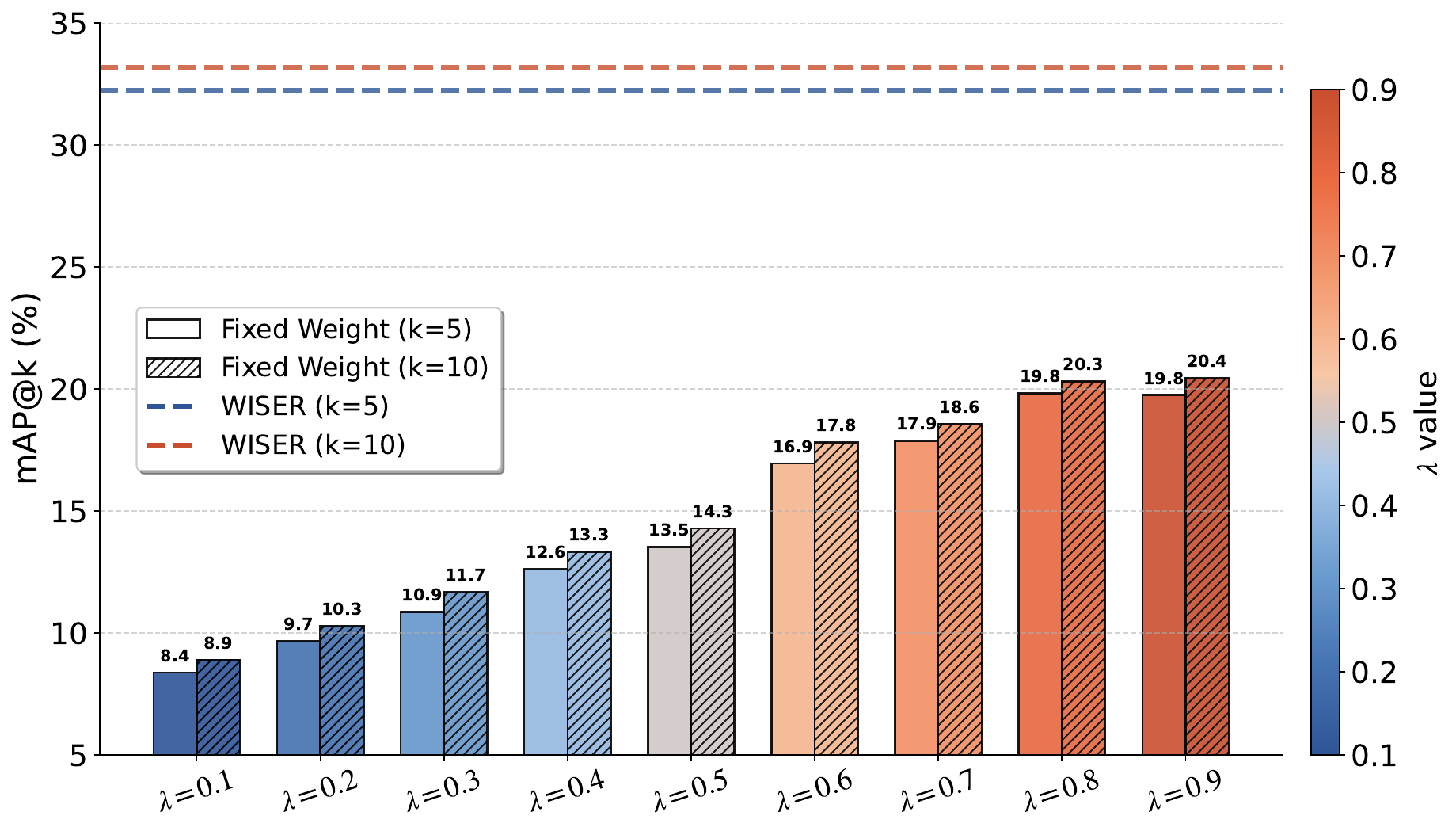}
   \caption{\textbf{Comparison between fixed fusion strategies and WISER on CIRCO.} $\lambda$ controls the T2I weight in the fixed fusion ($\lambda$ for T2I and 1-$\lambda$ for I2I). Our WISER method achieves superior performance over all $\lambda$ values, highlighting the limitation of static weighting.}
   \label{fig:fixed_weight}
   \vspace{-5pt}
\end{figure}

To further validate each component's contribution, we conduct additional ablations on the reranking component of Adaptive Fusion using single-path baselines (Table \ref{tab:more_ablation_fashioniq_circo}). Applied to the T2I-only baseline, it improves R@10 by 7.78\% on Fashion-IQ and mAP@5 by 11.12\% on CIRCO; for the I2I-only baseline, the gains are 9.99\% and 13.40\%, respectively. This shows that the verifier effectively identifies target candidates within each pathway's retrieval results. However, a significant gap remains compared to the full WISER framework, indicating that while reranking helps, the complementary strengths of T2I and I2I are essential for optimal performance in diverse CIR scenarios.

We further test other refiners (e.g, DeepSeek-R1, Gemini 2.5 Pro) in Table \ref{tab:further_ablation_circo_results}.
Both models achieve strong performance comparable to our default setup, confirming that WISER’s gains stem from the framework design rather than reliance on a specific model.
We empirically set the threshold to 0.7 and selected GPT-4o as the refiner without extra tuning. We believe targeted tuning could further improve performance, and thus report only the base results here.

% \begin{table}[h]
% \centering
% \caption{\textbf{More ablation study on the core components of WISER on the Fashion-IQ and CIRCO datasets.} ``AVG'' denotes average.}
% \resizebox{\linewidth}{!}{
% \begin{tabular}{cc|c|c|cc|cccc}
% \toprule
% \multicolumn{2}{c|}{\textbf{Wider Search}} & 
% \multirow{2}{*}{\makecell[c]{\textbf{Adaptive}\\\textbf{Fusion}}} & 
% \multirow{2}{*}{\makecell[c]{\textbf{Deeper}\\\textbf{Thinking}}} & 
% \multicolumn{2}{c|}{\textbf{Fashion-IQ-Avg}} &
% \multicolumn{4}{c}{\textbf{CIRCO}} \\
% \cmidrule(lr){1-2} \cmidrule(lr){5-6} \cmidrule(lr){7-10}
% \textbf{T2I} & \textbf{I2I} &  &  & \textbf{R@10} & \textbf{R@50} & \textbf{mAP@5} & \textbf{mAP@10} & \textbf{mAP@25} & \textbf{mAP@50} \\
% \midrule
% \checkmark & - & - & - & 28.59  & 49.18 & 17.28 & 17.94 & 19.64 & 20.51 \\
% \checkmark & - & \checkmark & - &  36.37 & 49.18 & 28.40 & 28.74 & 30.22 & 30.56 \\
% \checkmark & - & - & \checkmark & 29.22 & 49.94 & 17.64 & 18.30 & 20.01 & 20.94 \\
% - & \checkmark & - & - & 22.65  & 38.84 & 7.00 & 7.46 & 8.40 & 8.95 \\
% - & \checkmark & \checkmark & - & 32.64  & 38.84 & 20.40 & 20.12 & 20.63 & 20.73 \\
% - & \checkmark & - & \checkmark & 23.58 & 40.10 & 7.57 & 8.05 & 9.04 & 9.62 \\
% \checkmark & \checkmark & AVG & - & 33.40 & 52.92 & 13.53 & 14.30 & 15.96 & 16.77 \\
% \checkmark & \checkmark & \checkmark & - & 40.83  & 57.86 & 31.32 & 32.08 & 33.72 & 34.24 \\
% \checkmark & \checkmark & \checkmark & \checkmark & 41.99 & 58.74 & 32.23 & 33.18 & 34.82 & 35.35 \\
% \bottomrule
% \end{tabular}
% }
% \label{tab:more_ablation_fashioniq_circo}
% \end{table}
\begin{table}[h]
\centering
\caption{\textbf{More ablation study on the core components of WISER on the Fashion-IQ and CIRCO datasets.} ``RAK'' denotes rerank, ``AVG'' denotes average fusion at the similarity level, and ``ADA'' denotes Adaptive Fusion.}
% 调整行高：1.5 是拉伸系数，可以根据需要调整 (1.2, 1.8 等)
\renewcommand{\arraystretch}{1.2}
\resizebox{\linewidth}{!}{
\begin{tabular}{cc|c|c|cc|cccc}
\toprule
\multicolumn{2}{c|}{\textbf{Wider Search}} & 
\multirow{2}{*}{\makecell[c]{\textbf{Fusion}}} & 
\multirow{2}{*}{\makecell[c]{\textbf{Deeper}\\\textbf{Thinking}}} & 
\multicolumn{2}{c|}{\textbf{Fashion-IQ-Avg}} &
\multicolumn{4}{c}{\textbf{CIRCO}} \\
\cmidrule(lr){1-2} \cmidrule(lr){5-6} \cmidrule(lr){7-10}
\textbf{T2I} & \textbf{I2I} &  &  & \textbf{R@10} & \textbf{R@50} & \textbf{mAP@5} & \textbf{mAP@10} & \textbf{mAP@25} & \textbf{mAP@50} \\
\midrule
- & \checkmark & - & - & 22.65  & 38.84 & 7.00 & 7.46 & 8.40 & 8.95 \\
- & \checkmark & - & \checkmark & 23.58 & 40.10 & 7.57 & 8.05 & 9.04 & 9.62 \\
- & \checkmark & RAK & - & 32.64  & 38.84 & 20.40 & 20.12 & 20.63 & 20.73 \\
\checkmark & - & - & - & 28.59  & 49.18 & 17.28 & 17.94 & 19.64 & 20.51 \\
\checkmark & - & - & \checkmark & 29.22 & 49.94 & 17.64 & 18.30 & 20.01 & 20.94 \\
\checkmark & - & RAK & - &  36.37 & 49.18 & 28.40 & 28.74 & 30.22 & 30.56 \\
\checkmark & \checkmark & AVG & - & 33.40 & 52.92 & 13.53 & 14.30 & 15.96 & 16.77 \\
\checkmark & \checkmark & ADA & - & 40.83  & 57.86 & 31.32 & 32.08 & 33.72 & 34.24 \\
\checkmark & \checkmark & ADA & \checkmark & 41.99 & 58.74 & 32.23 & 33.18 & 34.82 & 35.35 \\
\bottomrule
\end{tabular}
}
\label{tab:more_ablation_fashioniq_circo}
\end{table}

\begin{table}[h]
\centering
\caption{\textbf{Further ablations of the refiner on CIRCO.}}
\renewcommand{\arraystretch}{1.0}
\resizebox{\linewidth}{!}{
\begin{tabular}{c|cccc}
\toprule
\multirow{1}{*}{\textbf{Method}} & 
\textbf{k=5} & \textbf{k=10} & \textbf{k=25} & \textbf{k=50} \\
\midrule
w/ Gemini 2.5 Pro & 32.92 & 33.65 & 35.4 & 35.95 \\
w/ DeepSeek-R1 & 33.08 & 33.85 & 35.51 & 36.05 \\
\midrule
\textbf{WISER (Ours)} & 32.23 & 33.18 & 34.82 & 35.35 \\
\bottomrule
\end{tabular}
}
\label{tab:further_ablation_circo_results}
\end{table}

\section{More Qualitative Examples}
\label{sec:more_qualitative}
In this section, we show more qualitative results on three datasets. 
We present the multimodal query, along with the edited image, edited caption, suggestions, and the top-5 retrieval results from T2I, I2I, and our method.

\subsection{More Qualitative results on CIRR}
\begin{figure*}[h]
  \centering
   \includegraphics[width=0.98\textwidth]{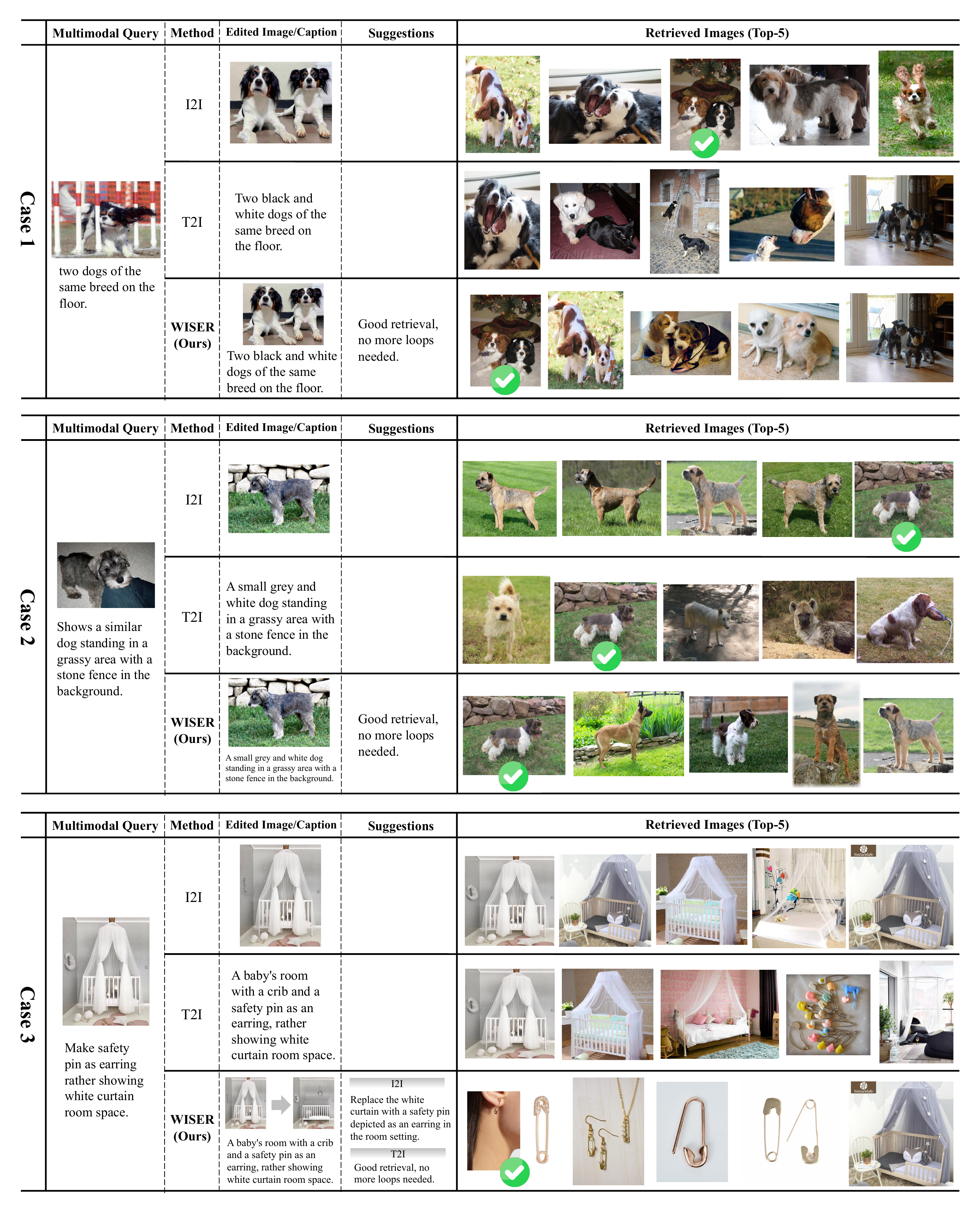}
   \caption{\textbf{More Qualitative Results on CIRR.} Green checkmark indicates correct target retrieval. WISER successfully retrieves the target image at top-1 across various scenarios.}
   \label{fig:suppl_cirr}
\end{figure*}

We present additional qualitative results on CIRR in Figure \ref{fig:suppl_cirr}.
In Case 1, WISER combines the strengths of I2I to capture fine-grained visual details (e.g., the dog’s breed) and T2I for semantic understanding, thereby retrieving the target image at top-1.
In Case 2, although I2I generates a reasonable edited image that follows the modification intent, the inherent fuzziness of ZS-CIR (e.g., variations in the dog’s orientation) introduces retrieval challenges. T2I retrieves more relevant candidates, benefiting from a more flexible textual representation. By adaptively fusing both pathways, WISER achieves top-1 retrieval of the target image.
Case 3 represents a highly complex modification where the target image has a weak correlation with the reference image. Both T2I and I2I baselines are misled by visual information from the reference, leading to initial retrieval failure. Consequently, this triggers Deeper Thinking to refine the edited image. Although some noise remains, WISER demonstrates strong robustness by successfully identifying the target image at top-1, highlighting its ability to handle abstract and semantically challenging edits.

\subsection{More Qualitative results on CIRCO}
\begin{figure*}[h]
  \centering
   \includegraphics[width=0.98\textwidth]{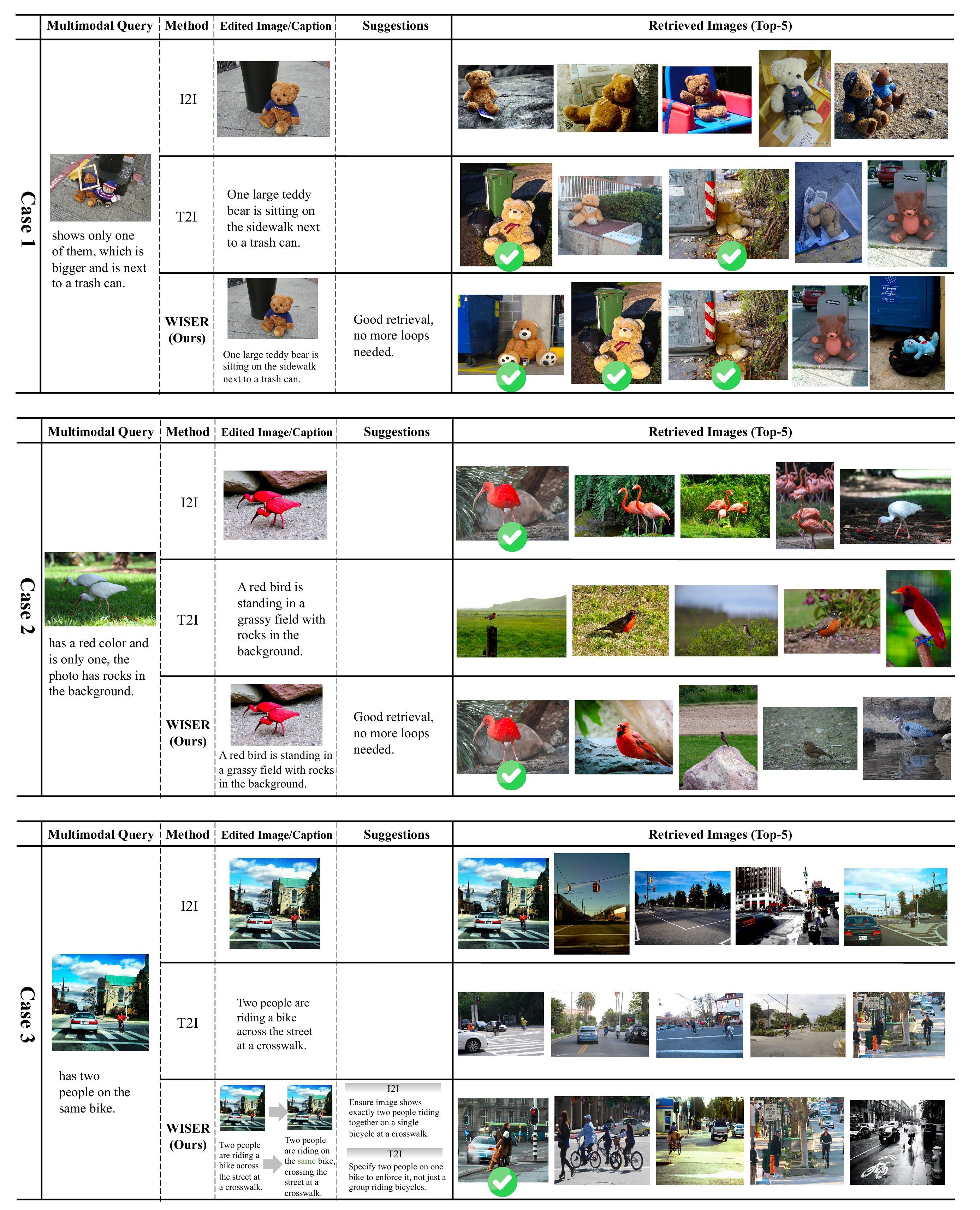}
   \caption{\textbf{More Qualitative Results on CIRCO.} Green checkmark indicates correct target retrieval. WISER successfully retrieves the target image at top-1 across various scenarios.}
   \label{fig:suppl_circo}
\end{figure*}

We present additional qualitative results on CIRCO in Figure \ref{fig:suppl_circo}. 
In Case 1, the modification intent is inherently ambiguous. I2I fails to retrieve the target, likely due to its strict reliance on visual similarity. In contrast, T2I successfully retrieves two target images within the top-5 by capturing key semantic elements while allowing for visual variation. WISER further expands retrieval diversity and returns more relevant targets through its adaptive fusion.
Case 2 demonstrates a scenario where visual precision is critical. I2I excels by preserving fine-grained details (e.g., the bird's breed) and successfully retrieves the target image at top-1. T2I suffers from the inherent ambiguity of textual representation and fails to identify the correct instance. WISER maintains the strong performance of I2I, highlighting its ability to preserve visual fidelity when it is essential.
Case 3 involves a complex compositional edit. Initially, both I2I and T2I struggle: I2I fails to generate a correct edited image (``two people on the same bike''), while T2I does not fully capture the precise semantic constraint (``on the same bike''). This uncertainty triggers Deeper Thinking. After refinement, both the edited image and caption accurately reflect the intended modification, enabling WISER to retrieve the target image at top-1 correctly. This case highlights the critical role of iterative refinement in resolving semantically and visually challenging queries.

\subsection{More Qualitative results on Fashion-IQ}
\begin{figure*}[h]
  \centering
   \includegraphics[width=0.98\textwidth]{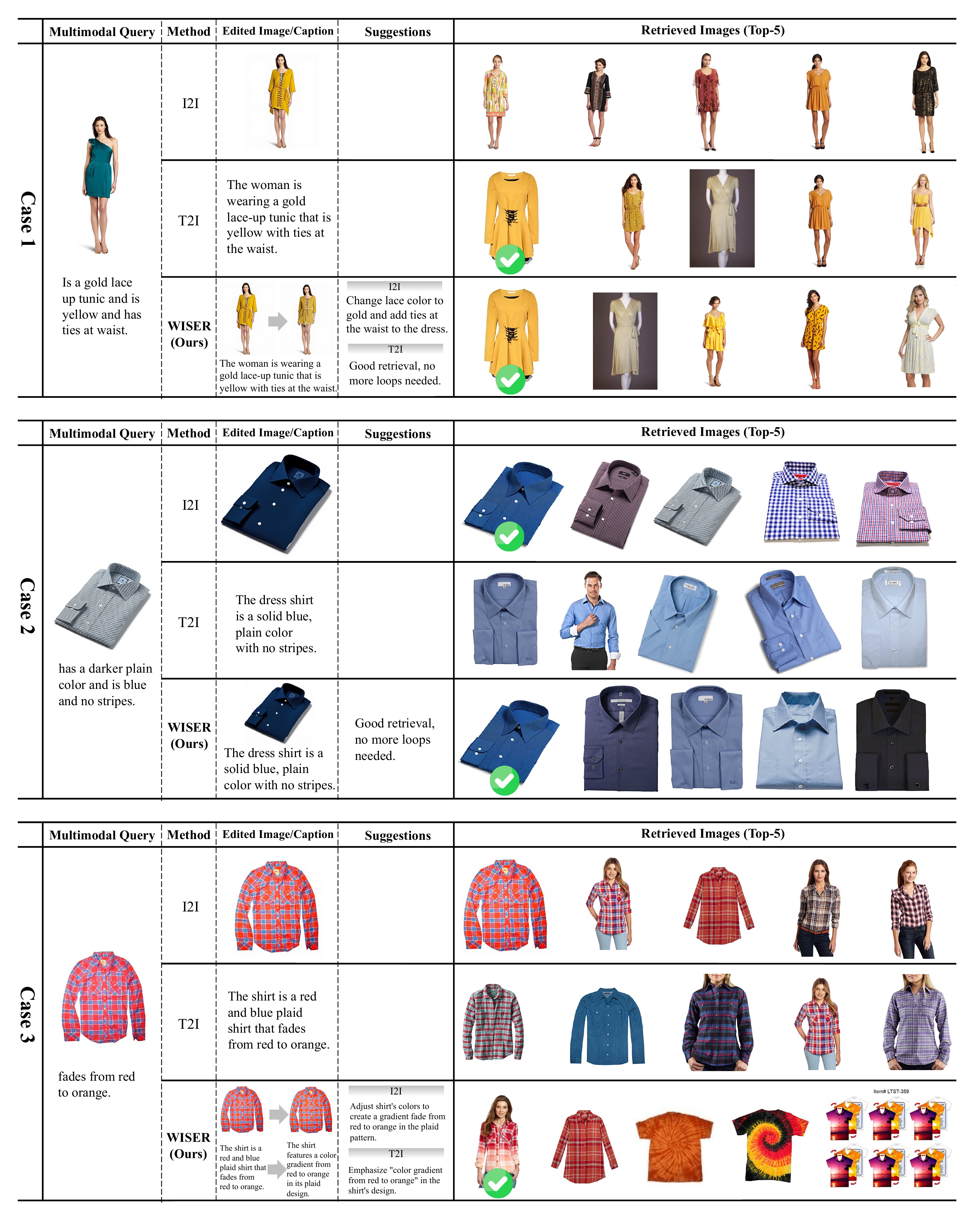}
   \caption{\textbf{More Qualitative Results on Fashion-IQ.} Green checkmark indicates correct target retrieval. WISER successfully retrieves the target image at top-1 across various scenarios.}
   \label{fig:suppl_fashioniq}
\end{figure*}

We present additional qualitative results on the Fashion-IQ dataset in Figure \ref{fig:suppl_fashioniq}.
In Case 1, due to the ambiguity in translating the specific attributes into a visual edit, I2I fails to retrieve the target accurately. In contrast, both T2I and WISER successfully retrieve the target image at top-1, demonstrating the advantage of semantic understanding in capturing detailed attribute-based changes.
In Case 2, I2I excels by preserving the structural details of the reference garment while accurately applying the color and pattern modifications, leading to correct top-1 retrieval. WISER maintains this strong performance through adaptive fusion.
Case 3 presents a more complex color transformation. I2I fails to generate a correct color gradient, while T2I introduces interference from the reference image by retaining the ``red and blue plaid'' pattern in its edited caption. This imprecision leads to retrieval inaccuracy. WISER, however, identifies the uncertainty and triggers Deeper Thinking to improve retrieval performance.

% \subsection{Pipeline Details}

\begin{figure}[h]
  \centering
   \includegraphics[scale=0.5]{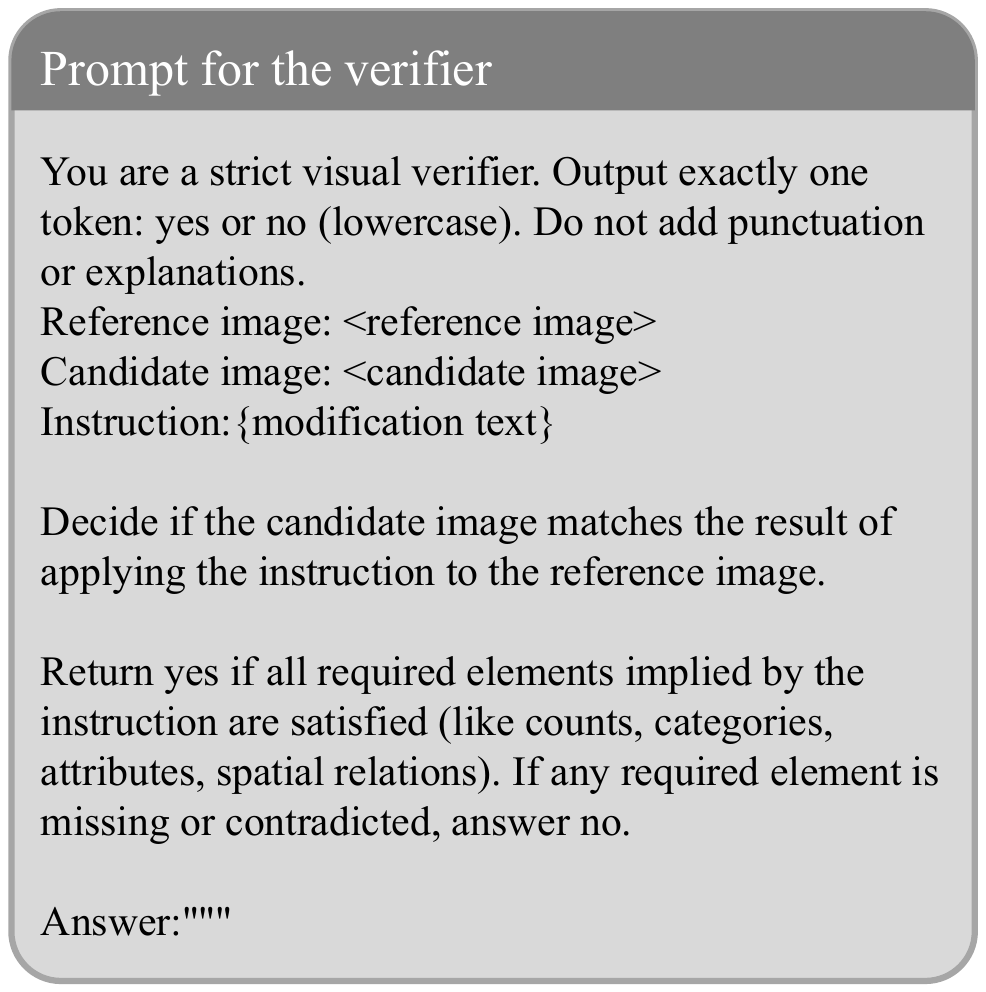}
   \caption{\textbf{Prompt for the verifier.} The prompt guides the verifier to answer a binary question given the reference image, the candidate image and the modification text.}
   \label{fig:prompt_verifier}
\end{figure}

\subsection{Failure Cases}
\begin{figure*}[h]
  \centering
   \includegraphics[width=0.98\textwidth]{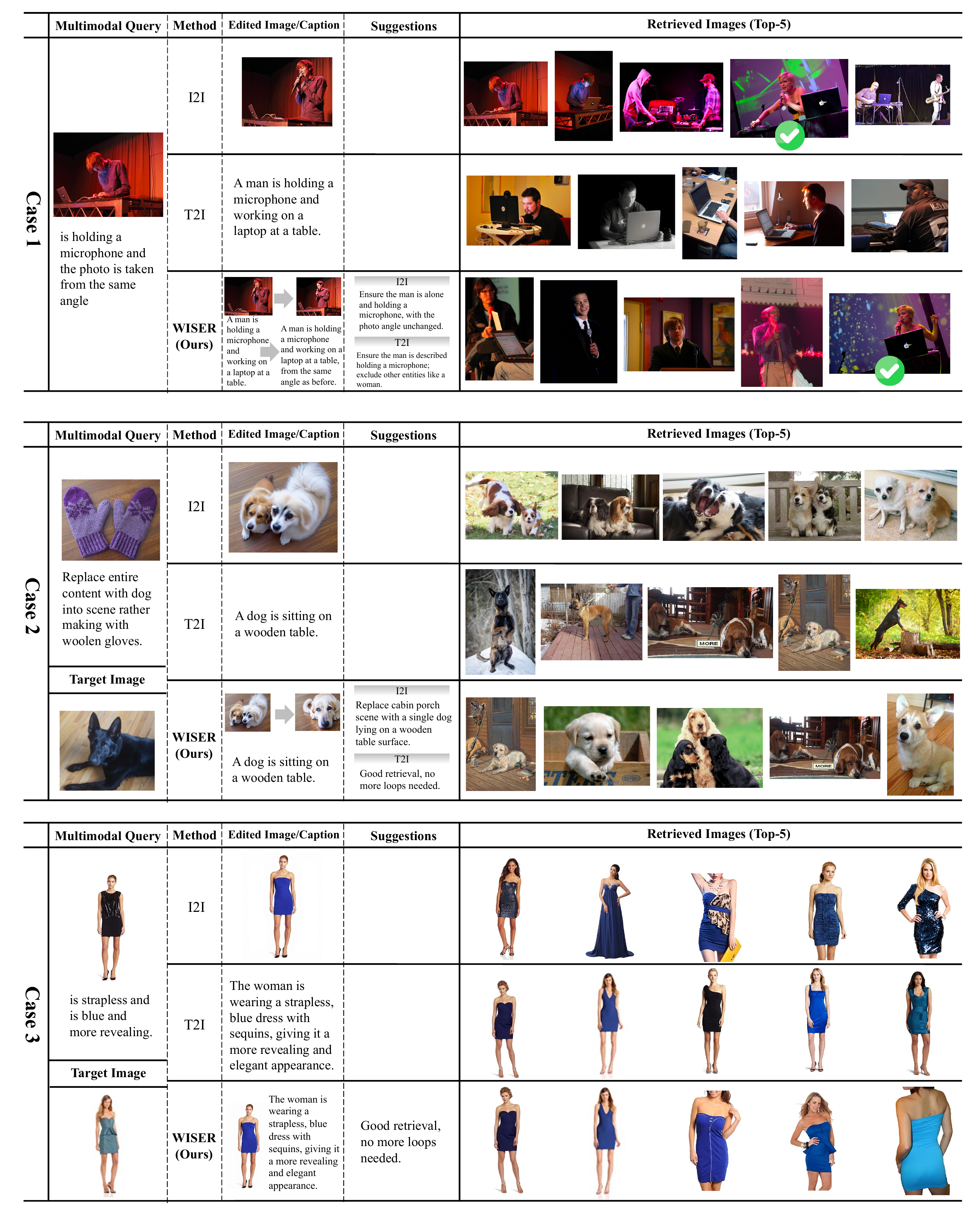}
   \caption{\textbf{Failure cases from CIRCO, CIRR, Fashion-IQ datasets.} We demonstrate three challenging cases where WISER struggles to retrieve the target image at top-1.}
   \label{fig:suppl_failure}
\end{figure*}

We also demonstrate failure cases across three datasets in Figure \ref{fig:suppl_failure}.
For Case 1 from CIRCO, I2I retains most visual information from the reference image and correctly generates the edited scene, which helps retrieve the target within top-5. However, it also introduces distraction by prioritizing stylistically similar but semantically unmatched images. T2I misunderstands the reference scene, incorrectly describing ``a man is working on a laptop'' instead of ``a DJ facing the camera with a console and a laptop.'' WISER overemphasizes the ``microphone'' attribute while neglecting other contextual information, leading to retrieval inaccuracy.
In Case 2 from CIRR, the modification requires complete replacement of the main subject, presenting a significant challenge. I2I mistakenly retains the entity count, generating two dogs instead of one. Although T2I captures the modification intent correctly, and WISER successfully refines the edited image after one iteration, all methods ultimately fail due to the high noise and inherent ambiguity in the CIRR dataset.
Case 3 from Fashion-IQ involves an ambiguous modification request. Due to the subjective nature of the description and the high visual similarity among fashion items, both T2I and WISER struggle to precisely identify the target from a large pool of candidate images with similar attributes.
This case underscores the difficulty in handling subjective or abstract attribute changes within a fine-grained retrieval domain.

% \begin{figure*}[h]
%   \centering
%    \includegraphics[width=0.9\textwidth]{pic/pipeline_details.pdf}
%    \caption{\textbf{Pipeline Details.}}
%    \label{fig:visualize}
% \end{figure*}

\section{Prompt}
\begin{figure}[h]
  \centering
   \includegraphics[scale=0.4]{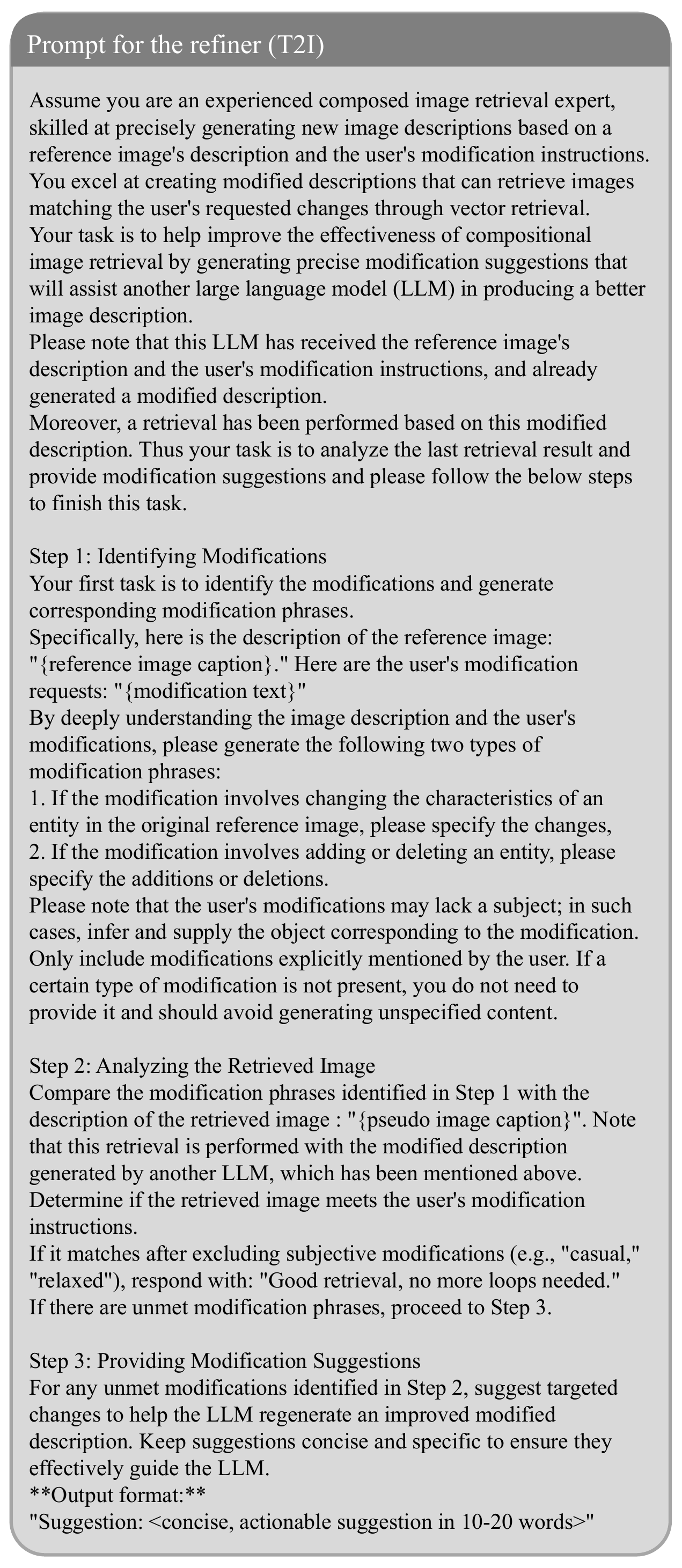}
   \caption{\textbf{Prompt for the refiner of T2I.}  The prompt guides the refiner to perform structured self-reflection for uncertain retrievals, given the reference image caption, modification text, and pseudo target caption from T2I. The generated suggestions are then fed to the editor to refine the edited caption for T2I.}
   \label{fig:prompt_refiner_t2i}
\end{figure}

\begin{figure}[h]
  \centering
   \includegraphics[scale=0.4]{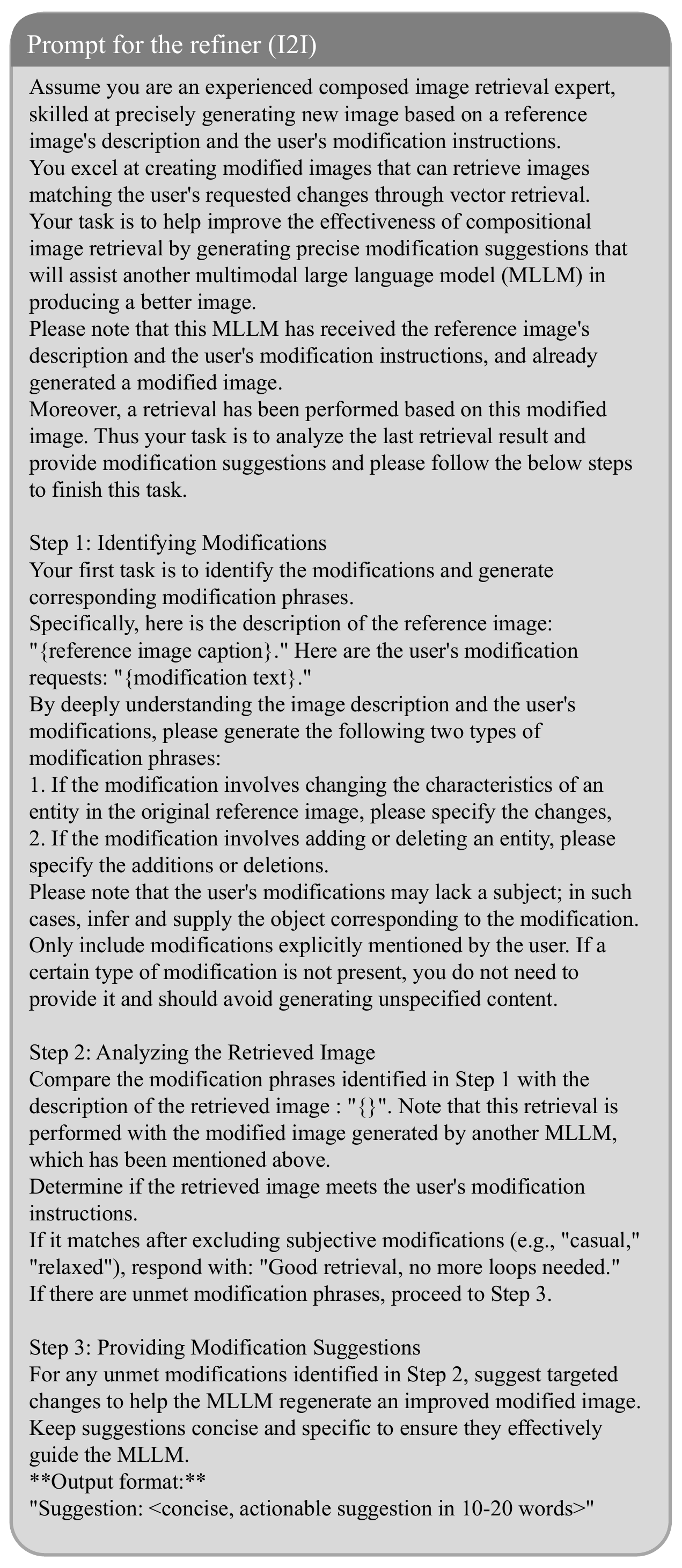}
   \caption{\textbf{Prompt for the refiner of I2I.} The prompt guides the refiner to perform structured self-reflection for uncertain retrievals, given the reference image caption, modification text, and pseudo target caption from I2I. The generated suggestions are then fed to the editor to refine the edited image for I2I.}
   \label{fig:prompt_refiner_i2i}
\end{figure}

In this section, we illustrate all the prompts used in our paper. 
% For Wider Search, we 参考了 CIReVL \cite{karthik2023vision}的方式 to generate edited caption and just feed modification text and reference image for image editing.
For Adaptive Fusion, we use the prompt shown in Figure \ref{fig:prompt_verifier}.
For Deeper Thinking, the prompt for T2I and I2I is shown in Figure \ref{fig:prompt_refiner_t2i} and Figure \ref{fig:prompt_refiner_i2i}, respectively.
Part of our prompts are taken from AutoCIR \cite{han2017automatic}.

\end{document}